\documentclass[a4paper,fleqn]{cas-dc}

\usepackage[numbers]{natbib}
\usepackage{subcaption}
\usepackage{float, placeins }

\def\tsc#1{\csdef{#1}{\textsc{\lowercase{#1}}\xspace}}
\tsc{WMH}
\tsc{OLL}
\tsc{MCC}
\tsc{CCE}

\begin{document}
\let\WriteBookmarks\relax
\def\floatpagepagefraction{1}
\def\textpagefraction{.001}

\shorttitle{Uncertainty Mapping in Fazekas Score Predictions}    

\shortauthors{S. Schmid et~al.}  

\title [mode = title]{Beyond Performance Metrics: Uncertainty Mapping of Label Ambiguity in Fazekas Score Prediction}  




\author[1,2]{Susanne Schmid}%[<options>]
%\credit{Conceptualization,Methodology, Software, Data curation, Formal analysis, Investigation, Validation, Visualization, Writing – original draft, Writing – review and editing.}

\author[2,3]{Johanna Ospel}
%\credit{Investigation, Validation, Writing – review and editing}
\author[2,4,6,1]{Richard Frayne}
\cormark[1]
%\credit{Resources, Data curation, Writing – review and editing}
\author[1,2]{Roberto Souza}%[]
%\credit{Funding acquisition,Project administration, Supervision, Writing – review and editing}

\cormark[1]


\cortext[1]{Corresponding and co-senior authors}
\ead{rfrayne@ucalgary.ca and roberto.souza2@ucalgary.ca}


\affiliation[1]{organization={Schulich School of Engineering, Department of Electrical and Software Engineering},
            addressline={University of Calgary}, 
            city={Calgary},
%          citysep={}, % Uncomment if no comma needed between city and postcode
            postcode={T2N 1N4}, 
            state={AB},
            country={Canada}}

\affiliation[2]{organization={Hotchkiss Brain Institute},
            addressline={University of Calgary}, 
            city={Calgary},
%          citysep={}, % Uncomment if no comma needed between city and postcode
            postcode={T2N 1N4}, 
            state={AB},
            country={Canada}}
\affiliation[3]{organization={Department of Radiology and Clinical Neurosciences},
            addressline={University of Calgary}, 
            city={Calgary},
%          citysep={}, % Uncomment if no comma needed between city and postcode
            postcode={T2N 1N4}, 
            state={AB},
            country={Canada}}

\affiliation[4]{organization={Calgary Image Processing and Analysis Centre},
            addressline={Foothills Medical Centre}, 
            city={Calgary},
%          citysep={}, % Uncomment if no comma needed between city and postcode
            postcode={T2N 2T9}, 
            state={AB},
            country={Canada}}
% \affiliation[5]{organization={Graduate Program in Biomedical Engineering},
%             addressline={University of Calgary}, 
%             city={Calgary},
% %          citysep={}, % Uncomment if no comma needed between city and postcode
%             postcode={T2N 1N4}, 
%             state={AB},
%             country={Canada}}

\affiliation[6]{organization={Seaman Family MR Research Centre},
            addressline={Foothills Medical Centre}, 
            city={Calgary},
%          citysep={}, % Uncomment if no comma needed between city and postcode
            postcode={T2N 2T9}, 
            state={AB},
            country={Canada}}

% Footnote of the second author
%\fnmark[]

% Email id of the second author
%\ead{}





\begin{abstract}
Reference labels used to train medical image classification models are not always as certain as they may appear, and this uncertainty has implications on performance metrics. In this study, we propose a framework to analyze model performance for periventricular Fazekas score prediction that goes beyond conventional metrics. The Fazekas score is an ordinal visual rating scale used to assess the severity of white matter hyperintensities and is known to be affected by inter-rater variability.  While the best Fazekas score prediction model achieved a Matthews correlation coefficient (MCC) of 0.70, performance varied across data splits and loss functions, making interpretation of model capabilities difficult.

Rather than interpreting epistemic uncertainty of a model's prediction as an isolated scalar value, our approach of uncertainty mapping relates uncertainty to its position within the learned feature representation. This highlights regions of class-boundary transitions where cases appear more ambiguous and misclassifications are more likely. It also identifies potential label disagreement, including low-uncertainty misclassified cases that expert review found to be inconsistent with the original reference Fazekas score. Therefore, uncertainty mapping allows model behaviour to be examined in relation to class separation and potential model-label disagreement.  
Loss function choice also influenced the uncertainty profile, with some models showing clearer class separation and more localized uncertainty in ambiguous regions than others. 
These findings suggest that uncertainty mapping for Fazekas score predictions can support model interpretation and targeted dataset review when reference labels are affected by ambiguity/ inter-rater variability. 
\end{abstract}


\begin{highlights}
\item Performance metrics alone did not fully capture model behaviour in the Fazekas score prediction task, which is affected by inter-rater variability. 
\item Mapping epistemic uncertainty in feature space helped distinguish ambiguous class-boundary cases from high-confidence misclassifications.
\item High-confidence misclassifications may indicate possible label inconsistency and can be prioritized for targeted expert review.
\item Noise-robust and ordinal losses produced models with different class separation capabilities and uncertainty patterns.
\item Clipped ordinal log loss improved class separation in the feature space compared with conventional classification losses. 

%\item 
\end{highlights}

%\nocite{*}

% Keywords
% Each keyword is separated by \sep

\maketitle

% Main text
\section{Introduction}
\label{introduction}
%Alternative Start:

White matter hyperintensities (WMH) are common brain lesions seen on T2-weighted fluid-attenuated inversion recovery (FLAIR) magnetic resonance imaging (MRI) that increase in severity with age. One widely used visual rating scale is the Fazekas score, an ordinal score assigned by radiologists based on lesion location and severity \cite{fazekas_mr_1987}. The score ranges from 0, indicating absence of WMH to 3, the most severe WMH appearance. Although Fazekas score is commonly applied to both periventricular and deep WMH regions, this study focuses on the prediction of periventricular Fazekas scores. Periventricular WMH are anatomically and pathologically heterogenous and with reported correlates including small vessel ischemia, demyelination, ependymitis granularis, and subependymal gliosis \cite{kim_classification_2008}. 

Despite the visual appearance of WMH on FLAIR images and the clinical relevance of Fazekas scores \cite{wardlaw_neuroimaging_2013}, automated severity assessment remains challenging. This raises an important question: why do machine learning models show limited performance in a task where the imaging findings are often visible and clinically meaningful?  One possible explanation is that the challenge does not lie only in the model, but in the way we assess performance. Conventionally, the performance of machine learning models is assessed using metrics such as accuracy, F1-score, and area under the curve of the receiver operating characteristic (AUROC), among others, estimating how well predictions align with the reference label. However, this performance-centered view may be incomplete because it assumes that the reference labels represent a stable and reliable ground truth. This assumption is ideal, but may not hold fully when the labels themselves are affected by a degree of variability, also referred to as label noise \cite{karimi_deep_2020, shi_survey_2024}.
In those scenarios, the search for an optimal model based on performance alone may overlook a more fundamental question: what do model errors reflect?

Fazekas scoring requires imaging findings to be assigned to a discrete ordinal category. As a result, cases near class boundaries may not map clearly onto a single score. Interpretation may depend on training and experience, and the presence of other lesions or imaging findings may influence the given Fazekas score \cite{gisev_interrater_2013, pesapane_errors_2024}.
This uncertainty is reflected in inter-rater variability, where different experts may assign different labels to the same case \cite{wardlaw_white_2004}. While consensus labels provide a practical reference standard, they do not necessarily remove the underlying ambiguity of the original ratings. Therefore, when such labels are used to train, validate, and compare machine learning models, conventional performance metrics may not necessarily reflect model capability alone, but may also reflect variability in the reference labels.

%When performance is limited, architectural modifications, changes in layer depth, hyperparameter tuning, and other adjustments are commonly performed to improve these metrics. 

This motivates the need to examine model behaviour beyond metrics. In this study, we propose uncertainty mapping, a framework that combines model uncertainty with learned feature-space representations to examine whether apparent model errors reflect model limitations, class ambiguity, or potential inconsistencies in the assigned Fazekas labels. 
Although epistemic uncertainty estimation and feature-space visualization are both established tools, the contribution of this work is to combine them so that epistemic uncertainty can be interpreted in relation to the model’s learned representation. 
Thereby, it becomes visible if a seemingly wrong prediction lies near the decision boundary, and carries therefore a degree of uncertainty; or if it falls within a region where labels seem to be certain and consistent.  

Using Fazekas score prediction as a case study, we examine model behaviour beyond standard performance metrics by applying uncertainty mapping to models trained with different loss functions, including noise-robust losses as well as losses incorporating the ordinal structure.

\section{Related Work} 
% add Fazekas score prediction work
The WMH severity and their progression are important markers of brain health, particularly in the context of cerebral small vessel diseases, healthy aging, and stroke \cite{wardlaw_neuroimaging_2013}. In large-scale studies, automated assessment of  WMH severity can support the inclusion of WMH burden in broader analyses without requiring extensive manual rating. Clinically, WMH severity is assessed using the Fazekas score, an established visual rating scale with meaningful associations to vascular brain health and clinical outcomes \cite{debette_clinical_2010, fazekas_mr_1987}. The Fazekas scale differentiates between periventricular and deep WMH, two lesion phenotypes that have been associated with partly distinct risk factors, clinical correlates, and pathological mechanisms \cite{griffanti_classification_2018, haller_brain_2013, krishnan_relationship_2006, lampe_visceral_2019, roseborough_microvessel_2022, sole-guardia_impact_2025, veldsman_spatial_2020}.

Much of the automated WMH literature has focused on lesion segmentation and volumetric quantification rather than direct prediction of Fazekas scores \cite{balakrishnan_automatic_2021,duarte_segmenting_2023, liu_deep_2020,umapathy_stacked_2021}.

Although WMH volume analysis is useful and provides a continuous measure of lesion burden, volumetric estimates can vary slightly across studies because of differences in MRI scanners, imaging protocols, and preprocessing pipelines \cite{bahrani_post-acquisition_2019,heinen_performance_2019, kruggel_impact_2010,melazzini_white_2021}. In addition, studies suggest that the clinical relevance of WMH depends not only on total lesion burden, but also on lesion location. For example, periventricular and deep WMH may show different associations with cognitive performance, and subclassification by location and signal intensity can reveal associations that are not captured by total WMH lesion volume alone \cite{bolandzadeh_association_2012, griffanti_classification_2018,haller_brain_2013, sole-guardia_impact_2025}. 
These findings support the continued relevance of clinically established visual severity scores, such as the Fazekas score, which encode regional and ordinal information.

Nevertheless, direct Fazekas score prediction remains challenging because the score converts gradual imaging patterns into discrete ordinal categories, making it sensitive to borderline cases and inter-rater variability \cite{haughey_assessment_2025, schmid_quantifying_2025, wardlaw_white_2004} .

Previous automated Fazekas scoring studies have used a range of evaluation strategies, including aggregated agreement metrics and binary comparisons between severity groups \cite{joo_diagnostic_2022, kuwabara_artificial_2024,rieu_fully_2023}. These metrics are useful for summarizing overall model performance; however, they provide limited insight into the actual Fazekas score model performance.
Philps \textit{et al.} \cite{philps_uncertainty_2025} supports the view that automated Fazekas score predictions is not only a model-performance problem. They showed that cross-dataset Fazekas classification can be affected by dataset-specific annotation criteria and differences in how borderline cases are scored. 

Fazekas score prediction illustrates a broader challenge in medical image classification: model behaviour can be difficult to interpret when reference labels contain ambiguity or noise. Since this study uses epistemic uncertainty \cite{gal_dropout_2016,kendall_what_2017} to examine model behaviour under label variability, we briefly review how uncertainty estimation has been used in related work. For example, uncertainty-aware learning approaches have used low-confidence predictions to reduce the influence of potentially noisy samples during training \cite{huang_uncertainty-aware_2022}. Xu \textit{et al.} \cite{xu_usdnl_2023} proposed a network that uses epistemic uncertainty to identify and select clean samples during training, thereby improving robustness to noisy labels.

Many proposed methods in the broader machine learning literature evaluate robustness to label noise using clean datasets with artificially introduced label corruption \cite{cordeiro_survey_2020, ghosh_making_2015,zhang_generalized_2018}. This is commonly done using symmetric noise, where labels are corrupted uniformly across classes, or asymmetric noise, where corruption is biased to specific classes. However, this type of artificial label noise can differ from label uncertainty typically encountered in the medical imaging domain. 
In medical image classification, label noise may arise not only from annotation errors, but also from expert disagreement \cite{karimi_deep_2020, song_learning_2023}. Such noise is often instance-dependent, meaning that the likelihood of disagreement or incorrect labeling depends on the characteristics of the individual image \cite{berthon_confidence_2021,frenay_classification_2014,karimi_deep_2020,menon_learning_2018,song_learning_2023,xia_part-dependent_2020}. 

Recent work has begun to address this issue more directly. In medical image classification, Ju \textit{et al.} \cite{ju_improving_2022} proposed a dual-uncertainty framework that uses epistemic and aleatoric uncertainty estimates to identify and re-weight potentially noisy labels during training.  Other approaches have focused on estimating cleaner label distributions by modelling instance-specific label confusion without requiring access to ground-truth labels \cite{liao_instance-dependent_2025}. 

Together, these studies show that label noise is increasingly understood as a sample-dependent problem rather than a simple random error. Yet, most existing approaches using uncertainty focus on improving training robustness or recovering cleaner labels. In contrast, this study uses uncertainty as an interpretive tool to examine where the model encounters ambiguity, how this ambiguity relates to class separation, and whether uncertain cases may indicate potential label disagreement.

\section{Methodology}
The task, dataset, and model architecture are built on our previous work on Fazekas score prediction \cite{schmid_quantifying_2025}, with the present study extending this framework with a clearer focus on uncertainty mapping, ordinal structure of the Fazekas score, and model optimization using task optimized loss functions. 

A machine learning model was trained to predict the periventricular Fazekas score, a visual rating scale used to assess WMH. The model was developed using the Calgary normative study (CNS) dataset \cite{mccreary_calgary_2020}, including 424 individuals ranging in age from 18 to 91 years. Participants had a median Montreal cognitive assessment (MoCA) score of $28\ (\text{IQR: }26\text{-}29)$ and sex was relatively balanced across the cohort (43.9\% male versus 56.1\% female). 

Volumetric 3D T1-weighted and 2D fluid-attenuated inversion recovery (FLAIR) images were acquired on a 3~T Magnetic Resonance (MR) scanner (Discovery MR750, General Electric (GE) Healthcare, Waukesha, WI). T1-weighted images were reconstructed at a resolution of $0.94 \times 0.94 \times 1.0$ mm$^3$. FLAIR Images were acquired axially at a resolution of $0.94 \times 0.94 \times 3.0$ mm$^3$.

\begin{figure}
\centering%% For centre alignment of image.
\includegraphics[width = 0.3\textwidth]{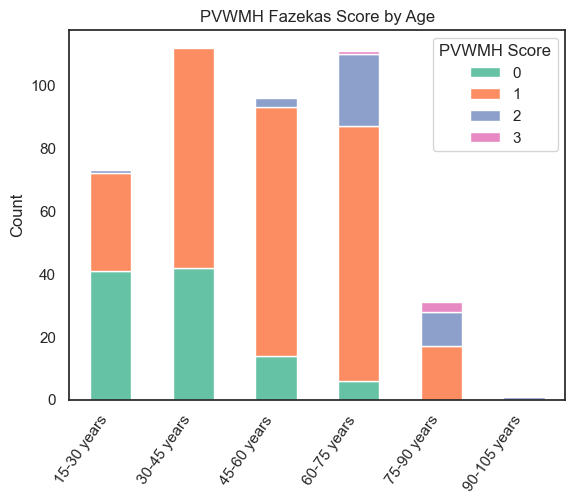}
%% Use \caption command for figure caption and label.
\caption{Distribution of the periventricular Fazekas score (PVWMH) by age. With increasing age, the WMHs become more prevalent. By the age of 80 years, all individuals have some extent of WMHs.}\label{pvwmh}
%% https://en.wikibooks.org/wiki/LaTeX/Importing_Graphics#Importing_external_graphics
\end{figure}

An experienced radiologist assessed WMH using the Fazekas scale \cite{fazekas_mr_1987}, a four-point visual rating scale ranging from 0 to 3, where 0 indicates the absence of WMH and 3 indicates the most severe burden. The scores were assigned for lesions in both deep white matter and periventricular areas (periventricular WMH), while this work focuses specifically on the periventricular Fazekas score. The distribution of periventricular Fazekas scores by age is shown in Figure \ref{pvwmh}.
Typical preprocessing steps for MR images were performed, including N4 bias field correction using ANTs \cite{avants_advanced_nodate,tustison_antsx_2021}, followed by image registration of FLAIR images to the corresponding T1-weighted volumes to ensure that both sequences resided in the same anatomical space. Brain extraction was performed on the T1-weighted scans using FSL-BET \cite{smith_advances_2004}, and the resulting brain masks were applied to the registered FLAIR images. Preceding, MR volumes were cropped to a dimension of 130x160x80 voxels to remove unnecessary background voxels. Image intensities were normalized using the Z-score of each volume.

\begin{figure}
	\includegraphics[width=\linewidth]{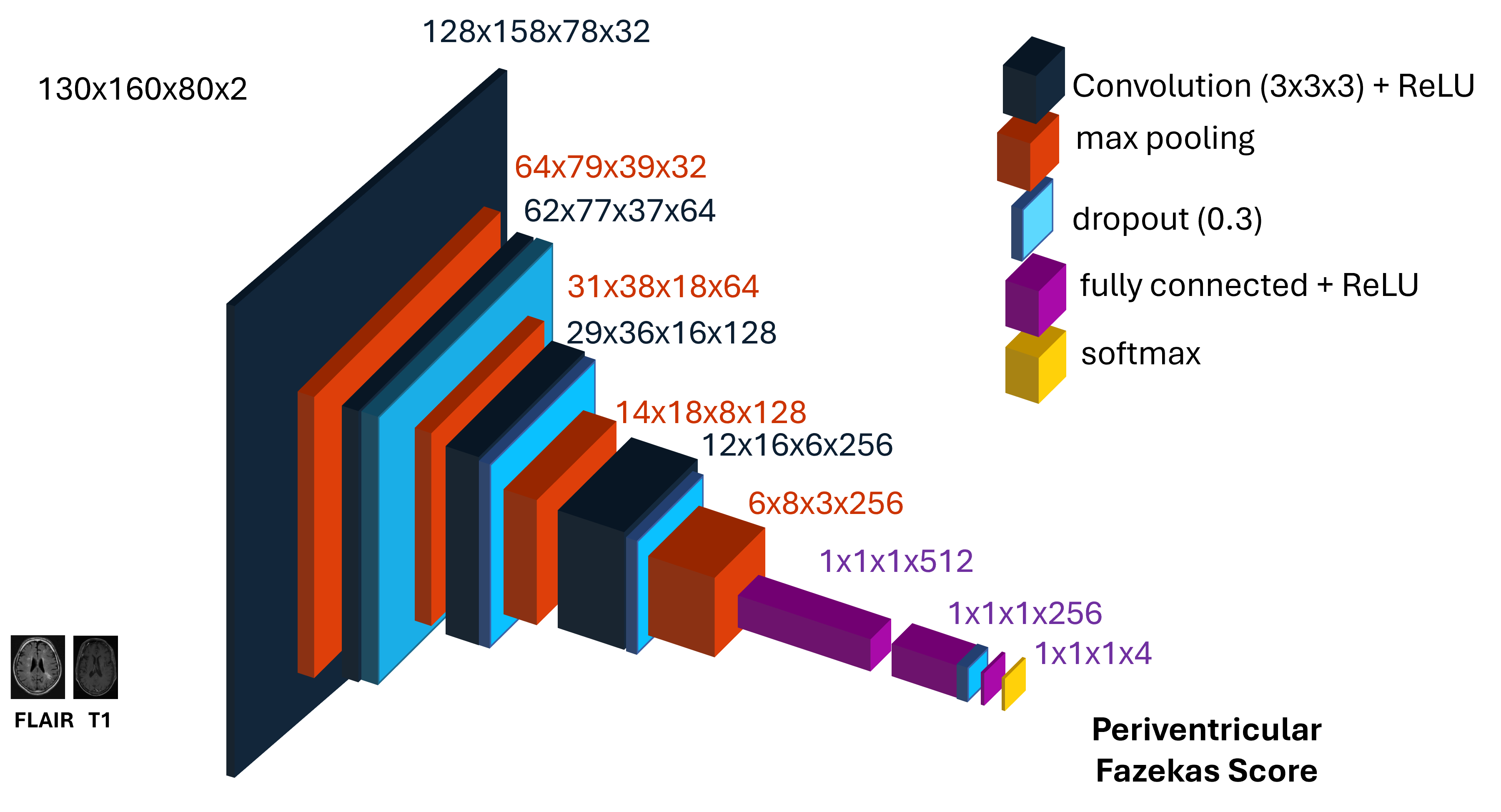}
	\caption{Two-channel 3D volume input to the convolutional neural network architecture. Multiple dropout layers were included to enable uncertainty estimation using Monte Carlo dropout.}
	\label{fig:MCD_model}
\end{figure}

A multi-channel 3D VGG-like convolutional neural network (Figure \ref{fig:MCD_model}) was trained to classify the periventricular Fazekas score using the preprocessed T1-weighted and FLAIR images as input. The model was optimized using AdamW with a weight decay of $4 \times 10^{-3}$, a learning rate of $7.7 \times 10^{-5}$, and a batch size of 16. Label smoothing of 0.2 was used as a regularization strategy. Due to the imbalanced distribution of Fazekas scores, soft stochastic class balancing was applied within each batch to increase the representation of minority classes during training. Data augmentation was also used to improve robustness, including random axial rotations of less than $10$ degrees, hemisphere flipping, and added Gaussian noise.

Four loss functions were evaluated to address classification-specific challenges, including label noise and ordinal class relationships. Categorical cross-entropy (CCE) loss was used as the standard classification baseline; however, it strongly penalizes predictions that assign a low probability to a specific class. When reference labels are ambiguous or potentially incorrect, this can encourage the model to fit uncertain labels too strongly. Based on this rational, we included generalized cross-entropy GCE, a label-noise-robust loss function \cite{zhang_generalized_2018} that has previously shown improved robustness under noisy-label settings. Ordinal log-loss (OLL) was included to account for the ordered (0,1,2,3) structure of the Fazekas score, penalizing predictions further away from the correct class more heavily. \cite{castagnos_simple_2022}. 
\begin{align}
   \mathcal{L}_{OLL}(p, y) =
-\sum_{i=1}^{C} \log(1 - p_i) \, d(y, i)^{\alpha} 
\end{align}
where $C$ is the number classes, $p_i$ the prediction output for class $i$, $y$ the reference class, and $d(y,i)$ the ordinal distance between class $i$ and class $y$. The parameter $\alpha$ controls the influence of the ordinal distance on the loss. For this study $\alpha = 1.5$.
In addition, we evaluated OLL with logit clipping (OLL clip). Following the LogitClip strategy \cite{wei_mitigating_2023}, the norm of the logit vector was clipped before the softmax using a threshold of $\tau$ = 0.9. If the logit vector exceeded this threshold, it was rescaled to have the norm $\tau$, while preserving its direction. The OLL loss was then computed from the clipped logits. Thus, OLL clip differs from OLL only in that the logits are norm-clipped before softmax. This was done to combine the ordinal structure of OLL with the noise-robustness motivation of logit clipping, which aims to bound the loss and so reduce the influence of large penalties arising from incorrect prediction, including those caused by noisy labels. 

The dataset was split into a nested cross-validation structure with five separate test sets, each containing four different training and validation set variations. Hyperparameters were fixed across all data splits. To reduce potential sampling bias, the splits were stratified by periventricular WMH severity.

Model performance was assessed using the Matthews correlation coefficient (MCC), which is well suited for evaluating classification performance in imbalanced and multi-class settings \cite{chicco_advantages_2020,chicco_matthews_2023}. MCC ranges from $-1$ to $+1$, where values near 0 indicate performance close to random prediction, and +1 indicates perfect agreement. In addition to MCC, model performance was assessed using accuracy, balanced accuracy, and per-class F1-score.

To better understand subject similarities, features from the penultimate layer (\textit{i.e.}, the second-to-last layer) of the model were extracted, and the 256-dimensional feature space was visualized using $t$-distributed stochastic neighbour embedding ($t$-SNE) for dimensionality reduction \cite{maaten_visualizing_2008}. The $t$-SNE implementation from the Scikit-learn library was used with perplexity = 13, n\_components = 2. All samples from the test are used for visualization.

\begin{figure*}
    \centering
    \includegraphics[width=0.9\linewidth]{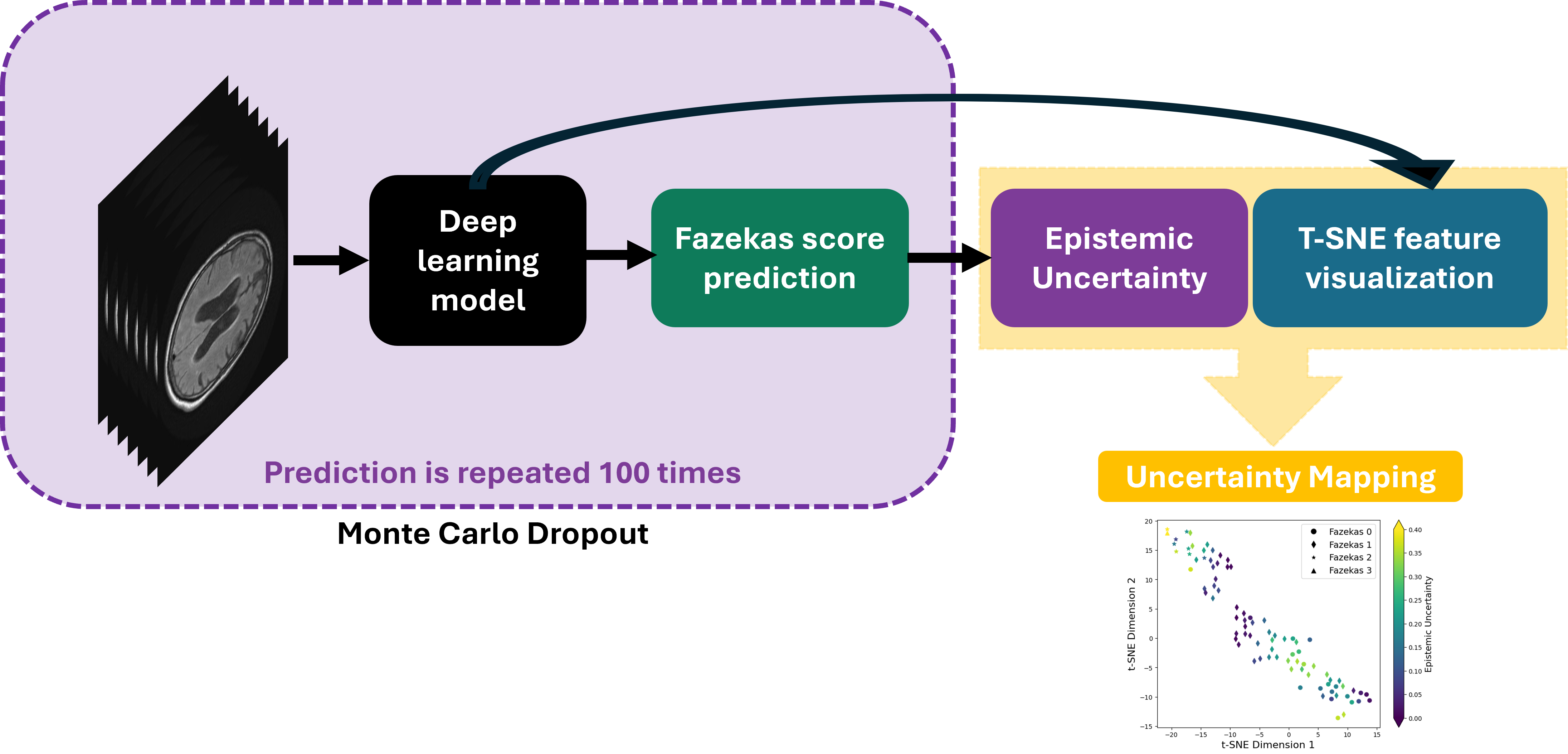}
    \caption{Uncertainty Mapping: Feature Visualization combined with epistemic uncertainty estimation obtained through Monte Carlo dropout.}
    \label{fig:UM}
\end{figure*}

Each test-set prediction was associated with uncertainty estimates derived from Monte Carlo dropout and visualized within the feature-space representation. This joint interpretation of feature similarity and uncertainty is referred to as uncertainty mapping (Figure \ref{fig:UM}).

To estimate model uncertainty, we used Monte Carlo dropout \cite{gal_dropout_2016}. During inference, dropout layers were kept active and each test sample was passed through the model 100 times, producing a distribution of predicted class probabilities. The final predictive distribution was obtained by averaging the predicted probabilities across all Monte Carlo samples.

We distinguish between predictive uncertainty and epistemic uncertainty. Predictive uncertainty $\mathbb{H}$ describes uncertainty in the final class prediction and was measured using the entropy of the averaged predictive distribution. Hence, the total predictive uncertainty $\mathbb{H}$ is defined as:
\begin{align} \label{eq:predUnc}
	\mathbb{H}[\bar{p}] = - \sum_{i=1}^C \bar{p}_i \log(\bar{p}_i)
\end{align}
while $\bar{p}_i$ are the mean predictions per class for all classes $C$.

The difference between total predictive uncertainty and expected uncertainty reflects epistemic uncertainty through mutual information as described in Equation \ref{eq:uncertainty}.
\begin{align} \label{eq:uncertainty}
	\text{Epistemic Uncertainty} = \underbrace{\mathbb{H}[\bar{p}]}_{\text{Total uncertainty}}  -  
	\underbrace{\mathbb{E}[\mathbb{H}[p_t]]}_{\text{Expected}}\\
    = - \left(\sum_{i=1}^C \bar{p}_i \log(\bar{p}_i) - \frac{1}{T} \sum_{t=0}^N  \sum_{i=1}^C p_{it} \log(p_{it})\right)
\end{align}
where $\mathbb{H}[\bar{p}]$ is the entropy of the average prediction (total uncertainty), $\mathbb{E}[\mathbb{H}[p_t]]$ the expected entropy (or aleatoric uncertainty), $C$ being number of classes, $N$ the number of stochastic forward passes and $p_{it}$ being a single prediction of the $t^{th}$ forward pass per class $i$.

%In addition, we report the standard deviation of the predicted class probabilities across Monte Carlo samples as a descriptive measure of prediction variability.

Model relevance maps were generated using a custom implementation of layer-wise relevance propagation (LRP) \cite{samek_evaluating_2017}, following the principles described by Montavon \textit{et al.} \cite{montavon_layer-wise_2019}.
Relevance was propagated backward from the predicted class score to the input image. In line with the recommended LRP settings, convolutional layers were propagated using the $\gamma$-rule with $\gamma = 0.01$, while fully connected layers were propagated using the standard LRP-0 rule. The $\epsilon$-rule was used for stabilization with $\epsilon = 0.01  \sigma(\text{image})$, where $\sigma(\text{image})$ denotes the standard deviation of the image intensities.

As a post-hoc analysis, targeted radiological re-evaluation was performed for a subset of samples identified from uncertainty mapping results. This analysis was not used for model training, selection, or performance metric calculation. Instead, it was intended to qualitatively assess whether uncertainty mapping could identify informative cases for expert review. The primary group consisted of all four low-uncertainty misclassified samples that were located within well-defined Fazekas score regions in the OLL clip representation. These cases were considered candidate examples of confident model-label disagreement.  Additionally to this type of group, two comparison groups were selected with four samples from regions of elevated uncertainty near class-boundary transitions, representing potentially ambiguous cases, and four low-uncertainty correctly classified samples located within well-defined Fazekas score regions, representing confident model-label agreement. All selected cases were reviewed by an experienced radiologist who was not involved in the original grading. The radiologist reassessed the periventricular Fazekas scores for each case and also indicated rating certainty. 

\begin{figure*}
    \centering
    \includegraphics[width=0.9\linewidth]{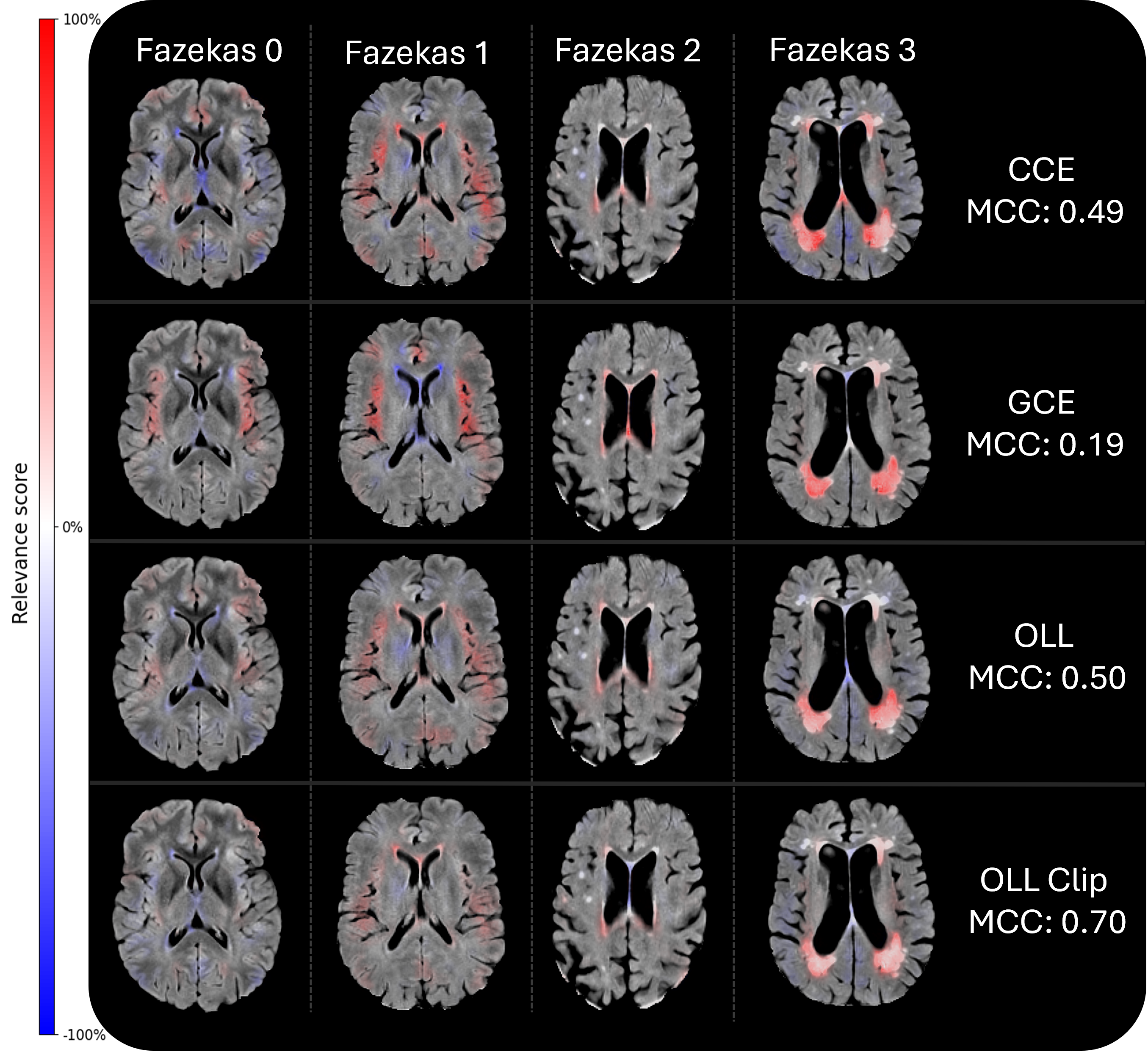}
    \caption{Visual comparison of Layerwise-Relevance Propagation (LRP) patterns for four representative subjects with periventricular Fazekas scores of 0, 1, 2, and 3. LRP heatmaps are shown for models trained with different loss functions from the fold with the best overall performance. While performance differed substantially between models, these differences were not clearly reflected in the heatmaps. Red regions indicate areas supporting the model prediction, whereas blue regions indicate areas contributing against the prediction or toward an alternative class. The subject with Fazekas score 1 was incorrectly predicted by the GCE-trained model.}
    \label{fig:LRP}
\end{figure*}

\begin{table*}[htbp]
	\caption{Nested cross-validation results: Average of all 20 runs $\pm$ the standard deviation.}
	\label{tab:nestedCrossVal_average}

	\begin{tabular}{lccccccc}
	\toprule
	Loss function & Accuracy & MCC & Cohen's $\kappa$ & F1: Class 0 & F1: Class 1 & F1: Class 2 & F1: Class 3 \\
	\midrule
	
	CCE 
	& 0.73 $\pm$ 0.05 
	& 0.46 $\pm$ 0.09 
	& 0.45 $\pm$ 0.09 
	& 0.54 $\pm$ 0.11 
	& 0.80 $\pm$ 0.05 
	& 0.65 $\pm$ 0.13 
	& 0.33 $\pm$ 0.44 \\
	
	GCE 
	& 0.68 $\pm$ 0.06 
	& 0.39 $\pm$ 0.07 
	& 0.37 $\pm$ 0.08 
	& 0.44 $\pm$ 0.19 
	& 0.76 $\pm$ 0.07 
	& 0.61 $\pm$ 0.14 
	& 0.24 $\pm$ 0.43 \\
	
	OLL 
	& 0.73 $\pm$ 0.04 
	& 0.47 $\pm$ 0.10 
	& 0.46 $\pm$ 0.11 
	& 0.53 $\pm$ 0.16 
	& 0.80 $\pm$ 0.04 
	& \textbf{0.66 $\pm$ 0.12} 
	& 0.31 $\pm$ 0.46 \\
	
	OLL clip 
	& \textbf{0.74 $\pm$ 0.05} 
	& \textbf{0.49 $\pm$ 0.10} 
	& \textbf{0.48 $\pm$ 0.10} 
	& \textbf{0.57 $\pm$ 0.10} 
	& \textbf{0.81 $\pm$ 0.04} 
	& \textbf{0.66 $\pm$ 0.14} 
	& \textbf{0.40 $\pm$ 0.46} \\
	
	\bottomrule
	\end{tabular}
\end{table*}

\section{Results}
Our deep learning approach predicted the periventricular Fazekas score with performance summarized in Table \ref{tab:nestedCrossVal_average}. More detailed results for each data split are provided in the Appendix Table \ref{tab:MCD_performance}. Performance varied across both data splits and loss functions used, despite using the same model architecture and fixed hyperparameters across all configurations.
Among the four loss functions analyzed, OLL clip showed a modest performance advantage, whereas GCE consistently showed weaker performance.

Despite these variations in performance, the relevance maps computed using LRP showed relatively consistent patterns across models. Relevance was mainly localized around periventricular WMH. In cases where WMH were absent or limited, relevance was observed in regions typically impacted by brain atrophy, an imaging marker commonly associated with WMH burden. These patterns suggest that the models generally relied on relevant anatomical image features for their prediction (Figure \ref{fig:LRP}). 

However, the heatmaps alone were insufficient to assess or explain differences in model performance, while the metrics revealed differences across loss functions and data splits. This suggests that differences in performance may not necessarily reflect major differences in the anatomical regions used by the models, but may instead relate to how each model learned from ambiguous or inconsistently labelled cases.

\subsection{Initial Uncertainty Analysis}
To further examine the observed performance differences, we focused on specific data splits. As a reference, we selected the split in which the CCE-trained model achieved its highest performance with an MCC of 0.59. On the same split, the models trained with OLL and OLL clip achieved slightly lower but comparable MCC values of 0.55, whereas the GCE-trained model showed a substantially lower MCC of 0.37. 

\begin{figure}
	\includegraphics[width=\linewidth]{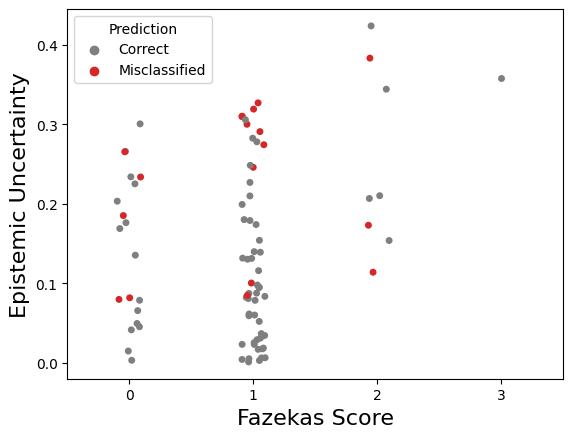}
	\caption{Distribution of epistemic uncertainty per Fazekas score class for the baseline model trained with CCE, achieving an MCC of 0.59. Misclassified cases are highlighted in red, while correct predictions are gray.}
	\label{fig:epi_dist}
\end{figure}

If inter-rater variability contributes to the observed performance differences, as hypothesized, samples with ambiguous or inconsistent labels might be expected to show elevated uncertainty. Figure \ref{fig:epi_dist} shows misclassified predictions of the CCE model and their corresponding epistemic uncertainty separated by Fazekas score. Although higher uncertainty was observed for some misclassified cases, increased uncertainty was not consistently associated with incorrect predictions. Several correctly classified samples also showed elevated uncertainty, while some misclassified samples were of relatively high confidence. These findings suggest that uncertainty alone does not explain the observed pattern of correct and incorrect predictions. 

\begin{figure*}[htb]
    \centering
    \includegraphics[width=0.7\linewidth]{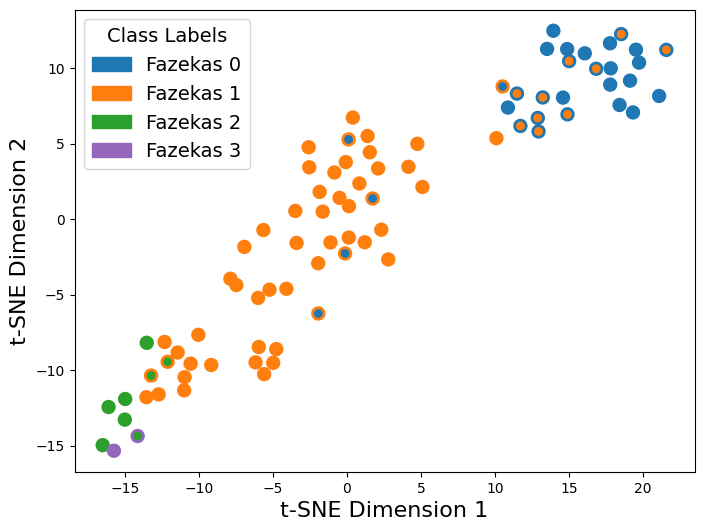}
    %  \vspace{1.5cm}
	\caption{$t$-SNE dimension reduction of the penultimate layer for CCE baseline model (MCC=0.59). Each point represents one subject from the test set. The fill colour indicated the reference label, while the edge colour indicated the model predictions. Hence, points with different fill and edge colours correspond to misclassified samples, while homogeneous points are correctly predicted. }
	\label{fig:t-SNE-results}
\end{figure*}

To further examine model behaviour under label ambiguity, we visualized the local relationship between test samples in the learned feature space. The resulting visualization is shown in Figure \ref{fig:t-SNE-results}.

An ordinal pattern across Fazekas score classes 0–3 is visible in the feature space, with samples generally progressing from lower to higher scores. However, for the reference labels, adjacent classes are not separated clearly, but rather show overlap. In contrast, the model predictions form more defined groups with clearer class boundaries. This pattern is expected because the feature representation is derived from the trained model and therefore reflects the internal structure used to generate the predictions \cite{salahuddin_transparency_2022}.

\begin{figure*}[htbp]
	\begin{minipage}[b]{0.49\linewidth}
		\centering
		\includegraphics[width=\linewidth]{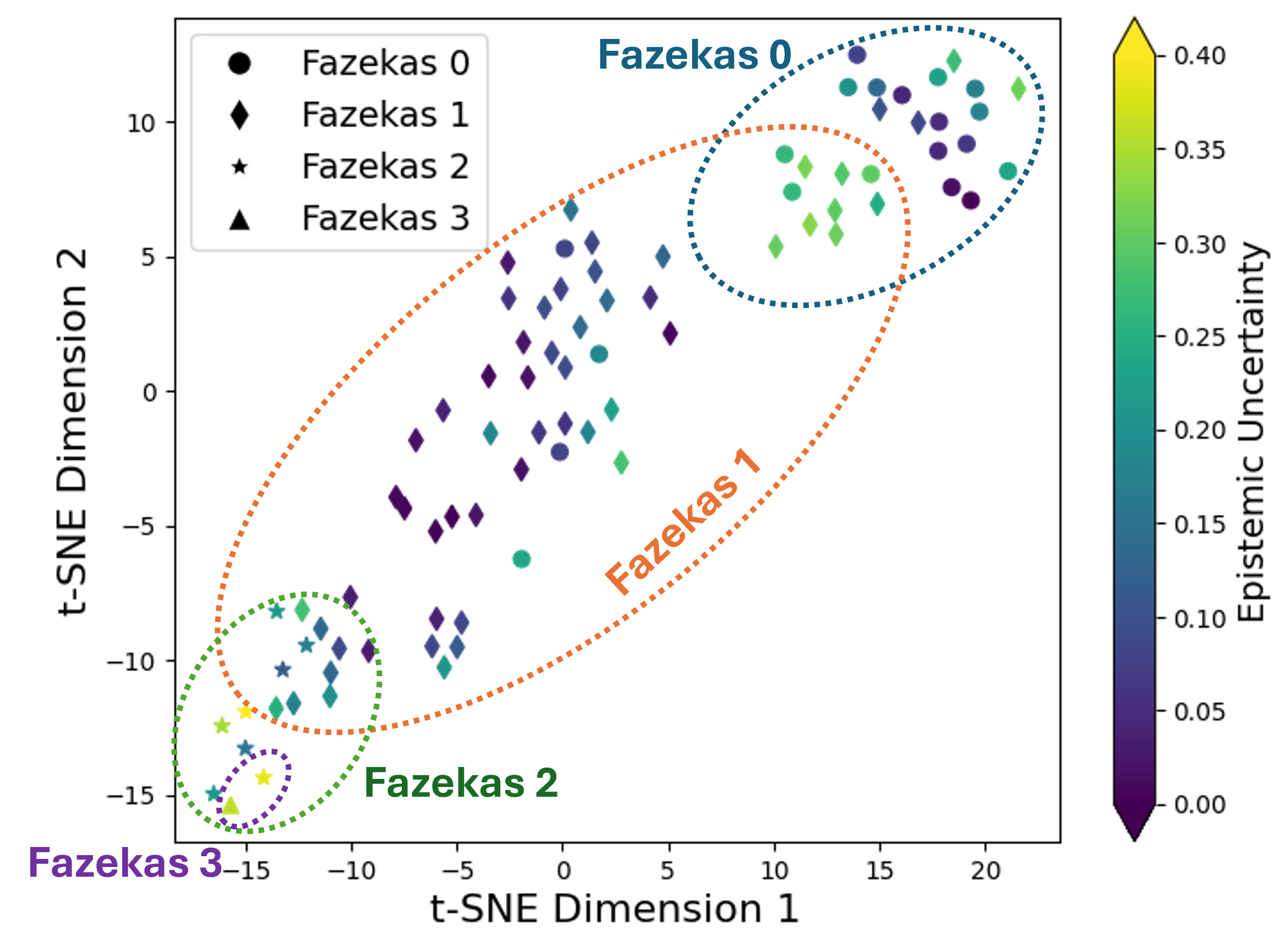}%MCD_test_t-SNE_epistemic_correct.png}
		\subcaption{Epistemic uncertainty}
		\label{fig:tSNE-uncertain-entropy}
	\end{minipage}
	\hfill
	\begin{minipage}[b]{0.49\linewidth}
		\centering
		\includegraphics[width=\linewidth]{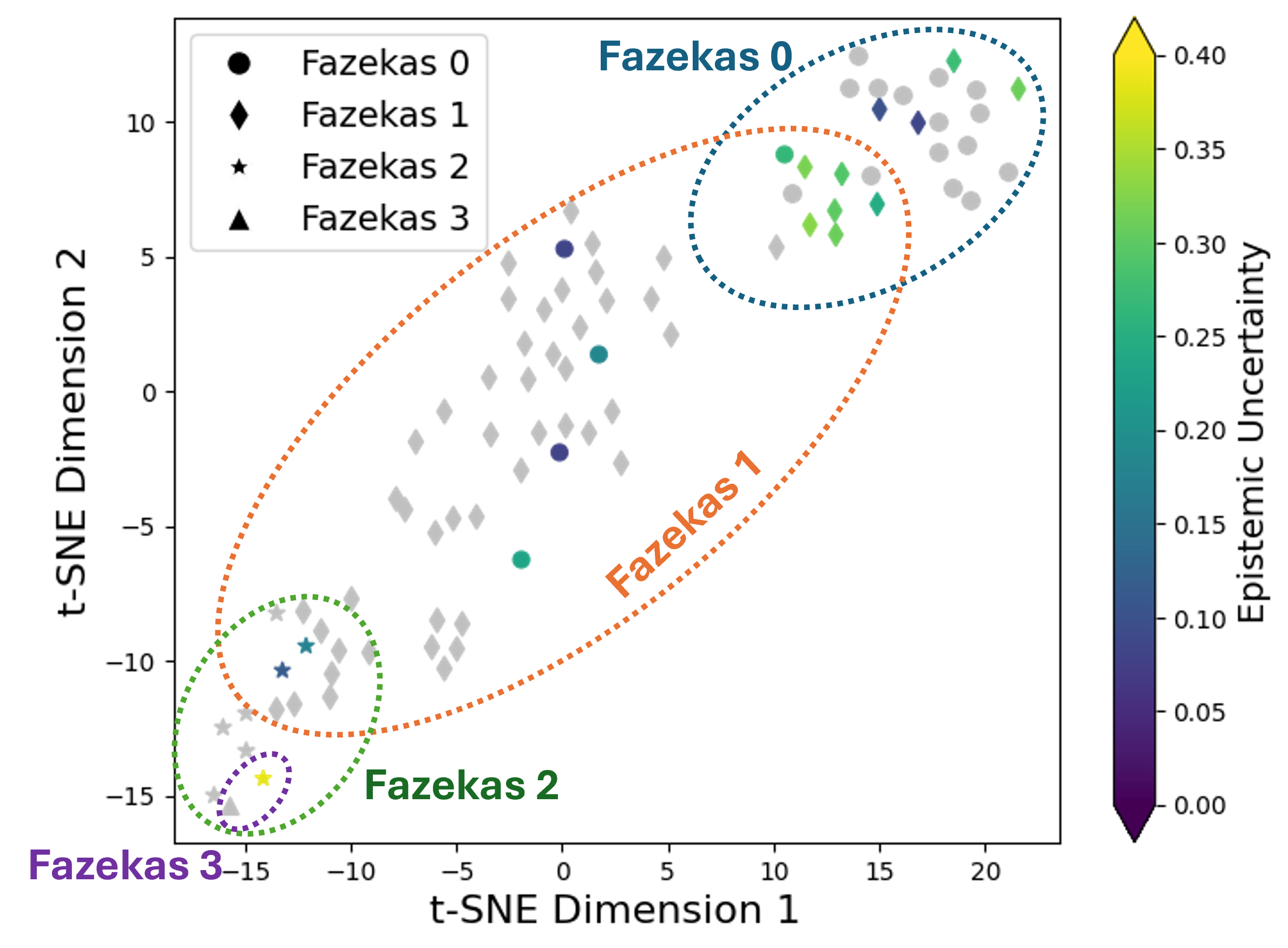}%MCD_test_t-SNE_epistemic_correct_agree.png}
		%  \vspace{1.5cm}
		\subcaption{Epistemic uncertainty of incorrect predictions.}
		\label{fig:tSNE-uncertain-std}
	\end{minipage}
	\caption{Uncertainty results combined with the $t$-SNE analysis. The shapes represent the reference Fazekas score, while the color scale represents epistemic uncertainty. In a), epistemic uncertainty is shown for all samples, while in b), the same results are visualized only for incorrect predictions; correct predictions are shown in grey. Dashed ellipses indicate Fazekas score region and were added to guide visual interpretation.}
	\label{fig:std-entropy-tsne}
\end{figure*}

To relate uncertainty to the model-specific feature-space representation, we overlaid the estimated epistemic uncertainty onto the t-SNE feature representation (Figure \ref{fig:std-entropy-tsne}). Uncertainty tended to be higher near class boundary transitions, \textit{e.g.}, between neighbouring Fazekas scores such as 0 and 1, or 1 and 2. In contrast, some misclassified samples showed relatively low uncertainty when located within more clearly defined class regions. 

Together, these findings suggest that uncertainty mapping provides a useful way to distinguish uncertain boundary cases from confident misclassifications in the learned feature space.

\subsection{Effect of Loss Function on Uncertainty Mapping}

To examine whether the observed uncertainty-mapping patterns were specific to the baseline split or were also present in another high-performing configuration, we analyzed the split in which the highest overall MCC was observed. In this split, the OLL clip model achieved the best performance, with an MCC of 0.70, followed by OLL, CCE, and GCE with MCC values of 0.50, 0.49, and 0.19, respectively. This split was used as an additional case to compare how the different loss functions were reflected in the learned feature representations and uncertainty patterns.

\begin{figure*}[htbp]
	\begin{minipage}[b]{0.47\linewidth}
		\centering
		\includegraphics[width=\linewidth]{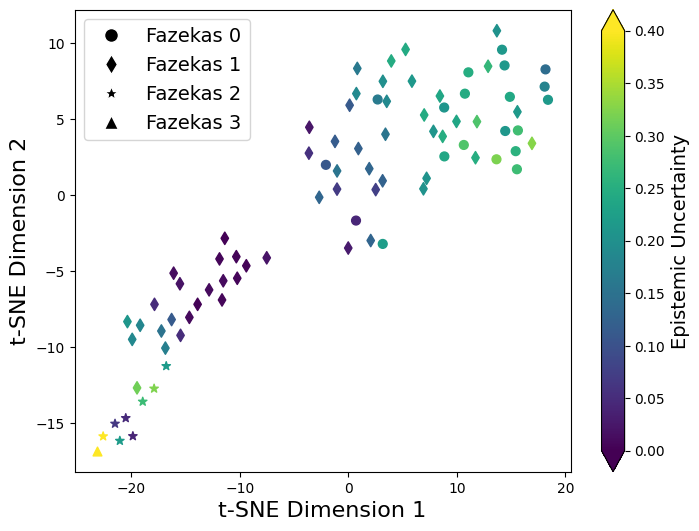}
        \subcaption{CCE - MCC: 0.49}
	\end{minipage}
	\hfill
	\begin{minipage}[b]{0.47\linewidth}
		\centering
		\includegraphics[width=\linewidth]{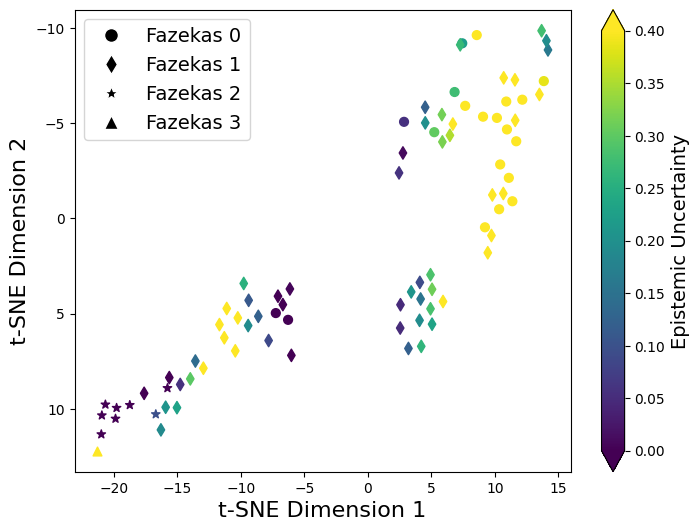}
        \subcaption{GCE -MCC: 0.19}
	\end{minipage}
    \vspace{0.5cm}
	\begin{minipage}[b]{0.47 \linewidth}
		\centering
		\includegraphics[width=\linewidth]{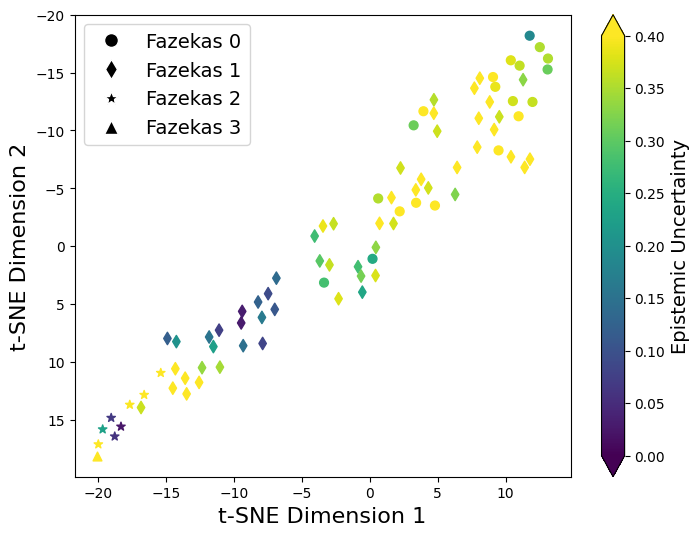}
        \subcaption{OLL - MCC: 0.50}
	\end{minipage}
    \hfill
	\begin{minipage}[b]{0.47 \linewidth}
		\centering
		\includegraphics[width=\linewidth]{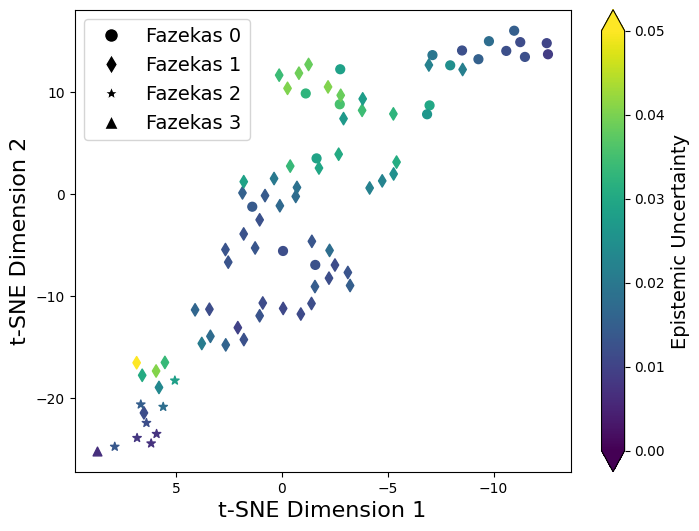}
        \subcaption{OLL clip - MCC 0.70}
	\end{minipage}
	\caption{Epistemic uncertainty estimation visualized combined with the corresponding model feature space representation for the four different loss functions analyzed (a) CCE, (b) GCE, (c) OLL, and (d) OLL clip. The color scale represents epistemic uncertainty, while the shape corresponds to the reference Fazekas score.}
	\label{fig:fold7-epistemic_correct-tsne}
\end{figure*}

The resulting t-SNE visualization showed noticeable differences in how the models organized test samples across loss functions (Figure \ref{fig:fold7-epistemic_correct-tsne}). While all models followed an ordinal trend consistent with the Fazekas score, OLL produced the clearest progression between neighbouring scores, in line with the ordinal structure incorporated into the loss function. Although CCE and OLL achieved similar quantitative performance, the OLL model exhibited higher epistemic uncertainty overall. Both CCE and OLL showed limited separation between Fazekas scores 0 and 1 in the learned feature space. In comparison, OLL clip had clearly defined regions of confident predictions across all labels, while increased epistemic uncertainty appeared more spatially concentrated near transitions between neighbouring classes. This suggests that OLL clip may have produced a representation in which uncertainty was more closely aligned with ambiguous ordinal boundaries. However, because logit clipping changes the scale and distribution of model confidence, absolute uncertainty values should be compared cautiously across clipped and non-clipped models. Overall, the main pattern was consistent with the baseline split. Uncertainty was often elevated near class-boundary transitions, but high uncertainty was not a consistent marker of misclassification. 

\begin{figure*}[htbp]
	\begin{subfigure}{0.47\linewidth}
	\includegraphics[width=\linewidth]{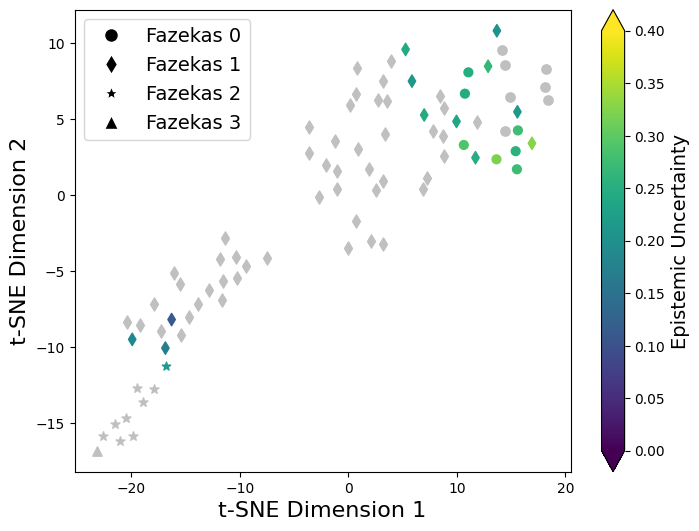}
	\subcaption{Uncertainty mapping of CCE}
	\end{subfigure}
	\begin{subfigure}{0.47\linewidth}
	\includegraphics[width=\linewidth]{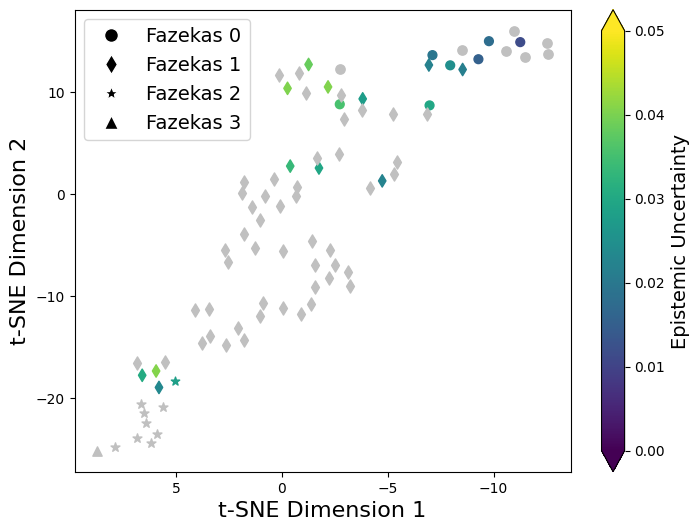}
	\subcaption{Uncertainty mapping of OLL clip }
	\end{subfigure}
	\caption{Agreement between the predictions of the CCE and OLL clip model is shown for both models uncertainty mapping. Coloured points indicate the epistemic uncertainty of test samples where the two models disagreed. Grey data points reflect subjects both models agreed on. The shape corresponds to the reference label.}
	\label{fig:UA-comparison}
\end{figure*}

This raised two further questions: to what extent do models agree on their predictions despite the difference observed in uncertainty mapping, and does model disagreement follow specific patterns within the learned feature space. 
To address this, we compared the predictions of the CCE and the OLL clip models using the split with the overall best performance. The two models showed only moderate agreement, with an MCC of 0.51. Figure \ref{fig:UA-comparison} shows the uncertainty mapping of the samples from which the two models disagreed. These disagreements were mainly located near transitions between neighbouring Fazekas scores, suggesting that the models differed most in regions where the classes were less clearly separable. However, these model disagreements did not necessarily correspond to samples that were misclassified relative to the reference label. 

\subsection{Targeted Label Re-evaluation}

\begin{figure}
    \centering
    \includegraphics[width=0.95\linewidth]{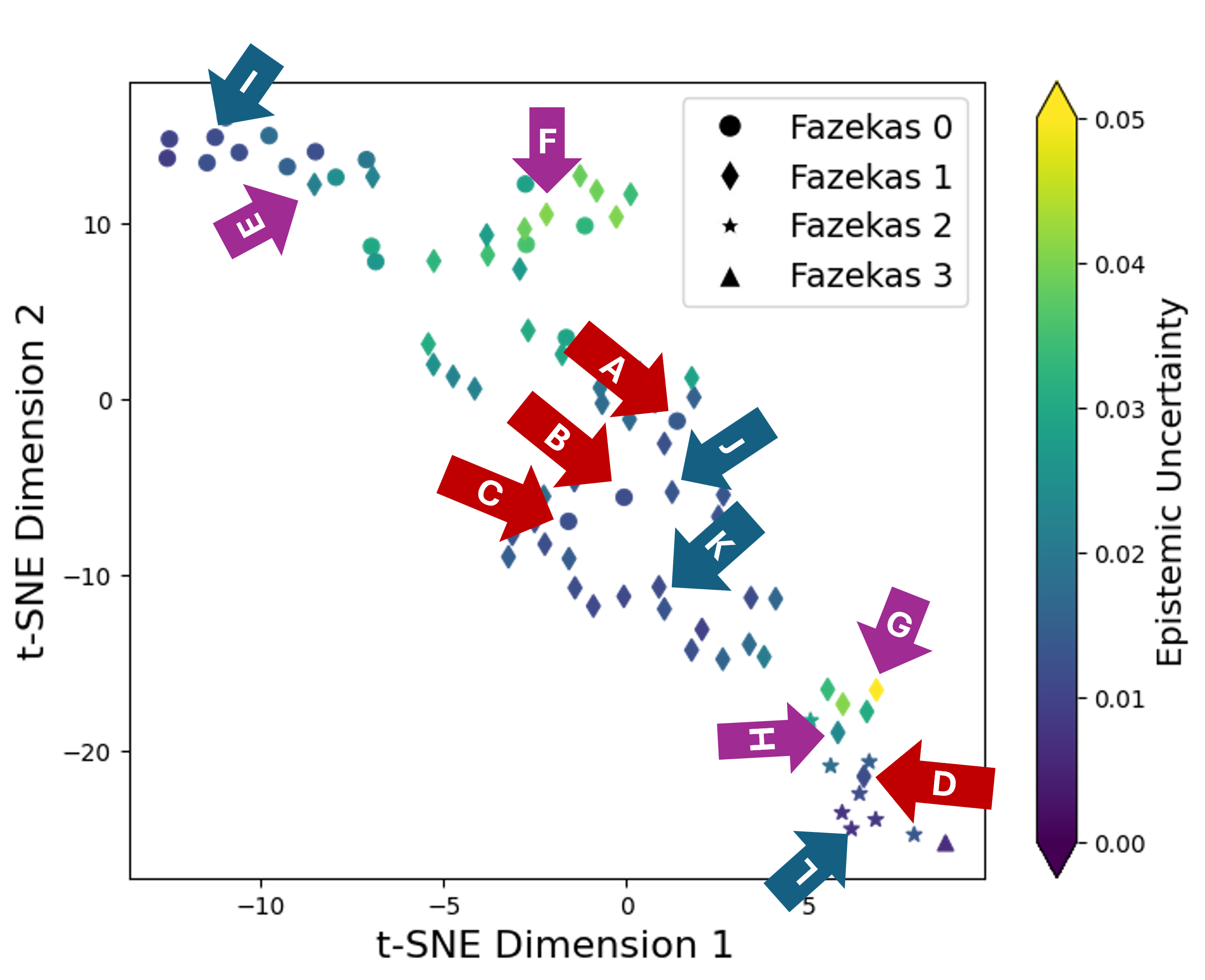}
    \caption{Identification of cases selected for targeted re-evaluation using the OLL clip uncertainty map. Three groups were selected: (1) label-inconsistency cases (A-D); (2) ambiguous boundary cases (E-H), selected from regions of elevated uncertainty near transitions between neighbouring Fazekas scores, and (3) confident agreement cases (I-L), defined as low-uncertainty correctly classified samples located within well-defined Fazekas score regions.}
    \label{fig:target_reevaluation}
\end{figure}

The best-performing model was used to determine targeted label re-evaluation. This model also showed a clearer class separation with elevated uncertainty concentrated near class boundary transitions. Based on the uncertainty map, four samples for each of the three groups (potential label inconsistency, ambiguous boundary region, and confident agreement) were selected (Figure  \ref{fig:target_reevaluation}). The corresponding MR FLAIR images of potential label inconsistency samples are shown in Figure \ref{fig:target_reevaluation}. In each example, the axial slice with the most prominent periventricular WMH was displayed. 
An independent radiologist not involved in the original grading re-assessed the selected subjects and reported the corresponding periventricular Fazekas score, as well as provided an indication of rating certainty for each reassessed case. The results are summarized in Table \ref{tab:target-reeval}.
Although the number of reviewed cases was small, the magnitude of label revision differed across the three groups. The label-inconsistency group showed the largest mean absolute revision with an average of 1.75 Fazekas points. In comparison, the ambiguous boundary and confident agreement groups showed smaller average changes of 0.75 and 0.25 Fazekas points, respectively. Together, those findings underline the presence of inter-rater variability across all classes, while the largest and most frequent revisions happened in the label-inconsistency group.

\begin{table*}[]
    \centering
    \caption{Target re-evaluation of the selected 12 cases summarized by group. Case letters correspond to the samples shown in Figure \ref{fig:target_reevaluation}. Label revisions are reported as original reference label $\rightarrow$ reassessed label. The symbols $+$ and $-$ indicate directional uncertainty in the reassessment score, where $+$ indicates that the score is at the higher end and $-$ indicates that the score is at the lower end. }
    \begin{tabular}{cccc}
        \toprule
         Selection Group & Cases Reviewed  & Revised & Observed revisions  \\
         \midrule
         Inconsistency& 4 & 4 & (A) 0$\rightarrow$2; (B) 0$\rightarrow$2; (C) 0$\rightarrow$1; (D)1$\rightarrow$3 \\
         Ambiguous boundary&4& 2 & (E) 1-; (H) 1$\rightarrow$2; (G) 1$\rightarrow$3-\\
         Confident &4&1& (I) 0$\rightarrow$1; (J) 1+ \\
         \bottomrule
    \end{tabular}
    \label{tab:target-reeval}
\end{table*}

%\begin{figure*}[htbp]
%	\begin{minipage}[b]{0.49 \linewidth}
%		\includegraphics[width=\linewidth]{figs/CCE_misclassfied_graph.png}
%		\subcaption{CCE model}
%		\label{fig:UA-missclassfiedGraphsCCE}
%	\end{minipage}
%	\begin{minipage}[b]{0.49 \linewidth}
%		\centering
%		\includegraphics[width=\linewidth]{figs/selcted_cases.png}
% 	\subcaption{OLL clip model}
%		\label{fig:UA-missclassfiedGraphs}	
	%\end{minipage}
%	\caption{Identification of four low-uncertainty misclassified cases selected for target re-evaluation. In the OLL clip model, the selected cases were located within regions of low epistemic uncertainty despite being misclassified relative to the reference label. When visualized using the CCE model, the same cases occupied different positions in the learned feature space. These four cases were subsequently reviewed by an experienced radiologist }
%   \label{fig:target_reevaluation}
%\end{figure*}

\begin{figure*}[htbp]
	\centering
	\includegraphics[width=1.\linewidth]{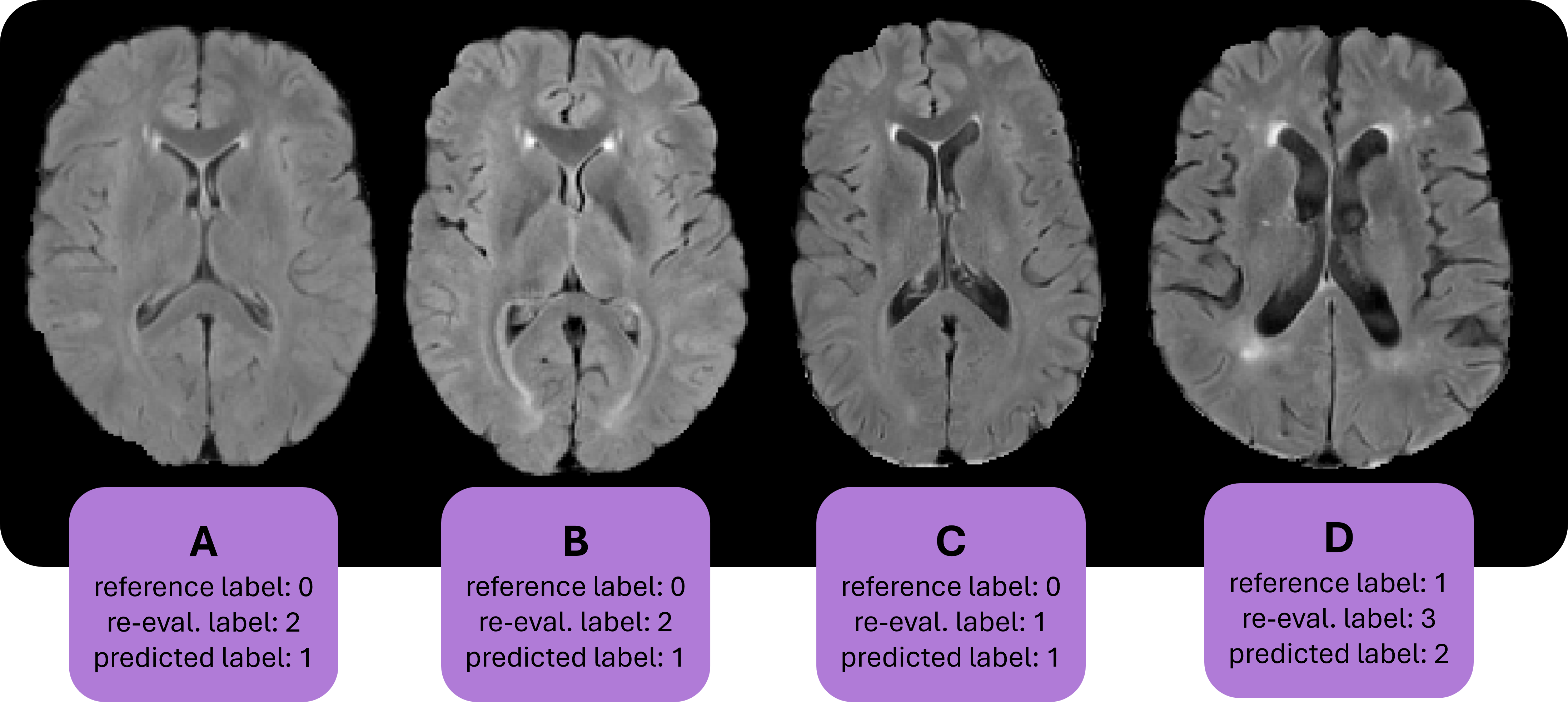}
	\caption{Label-inconsistency cases (misclassified despite low epistemic uncertainty) for the OLL clip model based on Figure \ref{fig:target_reevaluation}.}
	\label{fig:UA-missclassfiedImages}	
\end{figure*}
 
%\begin{figure*}
%	\centering
%		\includegraphics[width=1.\linewidth]{figs/OLLclip_correct_examples.png}
%	\caption{Correct predicted cases for the OLL clip model are shown for reference}
%	\label{fig:UA-correctImages}
%\end{figure*} 

\section{Discussion}\label{}

Deep learning models are increasingly used for medical image grading tasks, but their behaviour can be difficult to interpret when the reference labels contain ambiguity or variability. This study aimed to better understand this problem in the context of periventricular Fazekas score prediction from MR images. More specifically, we investigated how ambiguity in the reference labels may influence model interpretation. For this purpose, we proposed uncertainty mapping as an approach to visualize model uncertainty in the learned feature-space representation.

Although the overall model performance for periventricular Fazekas score prediction was moderate, it is useful to interpret these results within the context of the known label variability. Previous studies have reported inter-rater agreement for Fazekas scoring in the range of unweighted Cohen’s kappa ($\kappa$) values of approximately 0.34–0.42 \cite{haller_brain_2013}. This agreement can vary depending on factors such as lesion load, image quality, and rater experience \cite{haughey_assessment_2025,wardlaw_white_2004}. Some studies have reported higher agreement values; however, these are often based on weighted kappa estimates, which are not directly comparable to unweighted Cohen's kappa because of adjacent-category disagreements are penalized less severely. In this study, the models achieved unweighted Cohen’s kappa values of approximately 0.37–0.48 when compared with the reference labels and hence lie within the range of inter-rater agreement. Nevertheless, these results are not directly comparable because the models were trained on labels derived from the same grading process rather than acting as independent clinical rater.

The observed performance variability across data splits did not appear to reflect substantial differences in the image regions used for prediction, as indicated by the relatively consistent XAI heatmaps. Instead, the uncertainty mapping results in  Figure \ref{fig:UA-comparison} suggest that differences in model performance may be related to how cases are separated within the learned feature space. A possible explanation is that ambiguity in some reference labels due to inter-rater variability or borderline scoring decisions influenced the evaluation outcome. This interpretation was further supported by the target re-evaluation task, where expert review identified several uncertain or misclassified cases whose reference Fazekas score is potentially inconsistent.

The proposed uncertainty mapping approach provided this additional perspective on model behaviour and hence could serve as an interpretability technique. While feature-space representations have previously been used to inspect how deep learning models organize samples internally, they do not provide a complete explanation of individual predictions \cite{salahuddin_transparency_2022}. By adding epistemic uncertainty, we gain information how confident the model was within this organization. A model with clearer class separation and uncertainty concentrated near overlapping or boundary regions may be interpreted as more consistent with the ordinal and ambiguous nature of the grading task.

Beyond case-level uncertainty, the uncertainty maps also suggested differences in how clearly models organized the learned feature space. Some models showed more diffuse uncertainty patterns or weaker class separation, raising the question of whether less interpretable uncertainty structure is associated with poorer generalizability. Future work should therefore focus on whether uncertainty-map structure can be used not only to identify ambiguous cases, but also to compare model robustness across datasets and training strategies. 

Although GCE was included to improve robustness under potential label noise, it showed comparatively poor performance in this task while the LRP heatmaps still highlighted relevant image regions. This suggests that the type of label noise encountered in this task may differ from the noise settings in which GCE has commonly been evaluated. In many natural image classification benchmarks, label noise is simulated as symmetric or asymmetric label corruption, often with an assumed underlying ground truth \cite{zhang_generalized_2018}. In contrast, label variability in medical image grading is often instance-dependent, where anatomical variation, lesion burden, image quality, and borderline disease presentation can increase ambiguity in the assigned label \cite{xia_part-dependent_2020,liao_instance-dependent_2025}. Under these conditions, high-loss samples may not simply represent incorrect labels, but may also include clinically meaningful boundary cases. A noise-robust loss such as GCE may therefore reduce the influence of samples that are difficult but important for learning the ordinal structure of the task.

The ordinal structure of Fazekas scores is important when interpreting model performance, as adjacent predictions are less severe than larger score deviations. However, no clear metric improvement from CCE to OLL was observed. To further account for the noisy or ambiguous nature of the labels, we combined OLL with clipping. This combined loss appeared to improve predictive performance, suggesting that accounting for both ordinal structure and label uncertainty may be beneficial. 

Using uncertainty mapping revealed model differences despite comparable performance. While all loss functions reflected the ordinal nature of the Fazekas score to some extent, the clarity of class separation differed across models. OLL clip appeared to produce a more structured representation, with clearer separation between neighbouring Fazekas categories and only increased epistemic uncertainty near class boundaries.

During targeted re-evaluation, all four cases in the label-inconsistency group were revised relative to the reference label. Notably, three of these four cases differed by two Fazekas points after reassessment, indicating that the selected cases were not limited to minor boundary disagreements. Revisions were also observed in the ambiguity and confident group, although these were less frequent and smaller in magnitude. Overall, despite the small number of selected subjects, the re-evaluation supported the value of uncertainty mapping for identifying informative cases for expert review, while also highlighting the inherent ambiguity and inter-rater variability of Fazekas scoring. 

Several limitations should be considered. First, t-SNE was used to visualize the model's feature-space representation. However, it is an exploratory visualization method and does not preserve global distances. Therefore, interpretation was limited to local neighbourhood structure and qualitative patterns, rather than exact distances. Second, this study focused on epistemic uncertainty, which reflects uncertainty arising from the model itself. Incorporating, however, aleatoric uncertainty may provide a stronger perspective on label noise. Third, the analysis was based on the available reference labels rather than multi-rater consensus labels. Future work should investigate how uncertainty mapping behaves when consensus labels from multiple experts are available. Fourth, although the Fazekas score is widely used to grade WMH severity, it is not the only available visual rating scale, and different scales or annotation criteria may affect both model training and interpretation \cite{pantoni2002visual}. Finally, the present analysis was conducted within the available study cohort. In broader real-world clinical populations, co-existing pathologies such an old lacunar or territorial infarct, post traumatic encephalomalacia or other structural abnormalities may further complicate visual grading and could impact model performance. 

Overall, uncertainty mapping offers a complementary view of model learning beyond aggregate performance metrics. By combining feature-space representation with epistemic uncertainty, it provides a post-hoc approach for examining how label ambiguity may influence model interpretation. In this study, uncertainty mapping helped identify regions of class ambiguity near ordinal boundaries and low-uncertainty misclassified cases that may warrant targeted expert review, reducing the need for experts to re-evaluate the entire dataset. More broadly, the approach may also help flag samples where model behaviour is unexpected, which could motivate further investigation into possible model bias, limited generalizability, atypical imaging patterns, or even data quality issues. Future studies could investigate whether this approach can be used not only as an interpretability tool but also as part of an iterative model improvement strategy.


\section*{Data Availability}

The Calgary normative study data are not publicly available. The authors do not have authorization to redistribute the underlying imaging or clinical data.

\section*{Ethics Approval}
The study was approved by the Research Ethics Board at the University of Calgary (approval number 15-1285). All participants provided written informed consent. 

\section*{Acknowledgement}
The authors thank Dr. Amirmohammad Shamaei for helpful discussions regarding uncertainty estimation. The authors also thank the Calgary Normative Study investigators, research staff, imaging technologists, and participants for their contribution to the original data acquisition. 

\appendix
\section{Additional Model Results}

%\begin{table}[H]
\captionof*{table}{\textbf{Table A.1:} Model performance measured by Matthew-Correlation Coefficient (MCC) shown for the 20 models per loss function. Best performance per fold is highlighted in bold.}
\label{tab:MCD_performance}
\begin{tabular}{llrrrr}
	\toprule
	outer & inner & CCE & OLL & OLL clip & GCE \\
	\midrule
	0 & 0 & 0.41 & 0.43 & \textbf{0.60} & 0.34 \\
	& 1 & 0.42 & 0.44 & 0.43 & \textbf{0.46} \\
	& 2 & 0.41 & 0.47 & 0.43 & \textbf{0.48} \\
	& 3 & 0.55 & \textbf{0.60} & 0.51 & 0.43 \\
	\midrule
	1 & 0 & 0.41 & \textbf{0.55} & 0.46 & 0.43 \\
	& 1 & 0.55 & 0.57 & \textbf{0.63} & 0.38 \\
	& 2 & 0.51 & 0.38 & \textbf{0.56} & 0.34 \\
	& 3 & 0.49 & 0.50 & \textbf{0.70} & 0.19 \\
	\midrule
	2 & 0 & 0.55 & 0.55 & \textbf{0.59} & 0.49 \\
	& 1 & 0.46 & \textbf{0.58} & 0.48 & 0.46 \\
	& 2 & 0.47 & 0.51 & \textbf{0.57} & 0.49 \\
	& 3 & \textbf{0.59} & 0.55 & 0.55 & 0.37 \\
	\midrule
	3 & 0 & 0.35 & 0.29 & 0.30 & \textbf{0.45} \\
	& 1 & 0.22 & 0.23 & \textbf{0.32} & \textbf{0.32} \\
	& 2 & 0.48 & \textbf{0.53} & 0.48 & 0.23 \\
	& 3 & \textbf{0.56} & 0.46 & 0.40 & 0.41 \\
	\midrule
	4 & 0 & 0.30 & 0.46 & \textbf{0.53} & 0.30 \\
	& 1 & 0.32 & 0.41 & \textbf{0.42} & 0.25 \\
	& 2 & \textbf{0.47} & \textbf{0.47} & 0.44 & 0.37 \\
	& 3 & 0.39 & 0.42 & \textbf{0.46} & 0.37 \\
	\bottomrule
	\end{tabular}

%\end{table}

%\section{}\label{}

\bibliographystyle{cas-model2-names}

%\bibliography{cas-refs}
\bibliography{ref}

@article{pantoni2002visual,
  title={Visual rating scales for age-related white matter changes (leukoaraiosis) can the heterogeneity be reduced?},
  author={Pantoni, Leonardo and Simoni, Michela and Pracucci, Giovanni and Schmidt, Reinhold and Barkhof, Frederik and Inzitari, Domenico},
  journal={Stroke},
  doi={10.1161/01.STR.0000038424.70926.5E},
  volume={33},
  number={12},
  pages={2827--2833},
  year={2002},
  publisher={Lippincott Williams \& Wilkins}
}

@article{menon_learning_2018,
	title = {Learning from binary labels with instance-dependent noise},
	volume = {107},
	issn = {1573-0565},
	doi = {10.1007/s10994-018-5715-3},
	pages = {1561--1595},
	number = {8},
	journaltitle = {Machine Learning},
	shortjournal = {Mach Learn},
	author = {Menon, Aditya Krishna and van Rooyen, Brendan and Natarajan, Nagarajan},
	date = {2018-09-01},
	langid = {english},
}

@article{pesapane_errors_2024,
	title = {Errors in Radiology: A Standard Review},
	volume = {13},
	rights = {http://creativecommons.org/licenses/by/3.0/},
	issn = {2077-0383},
	doi = {10.3390/jcm13154306},
	shorttitle = {Errors in Radiology},
	pages = {4306},
	number = {15},
	journaltitle = {Journal of Clinical Medicine},
	publisher = {Multidisciplinary Digital Publishing Institute},
	author = {Pesapane, Filippo and Gnocchi, Giulia and Quarrella, Cettina and Sorce, Adriana and Nicosia, Luca and Mariano, Luciano and Bozzini, Anna Carla and Marinucci, Irene and Priolo, Francesca and Abbate, Francesca and Carrafiello, Gianpaolo and Cassano, Enrico},
	date = {2024-01},
	langid = {english},
}

@online{cordeiro_survey_2020,
	title = {A Survey on Deep Learning with Noisy Labels: How to train your model when you cannot trust on the annotations?},
	shorttitle = {A Survey on Deep Learning with Noisy Labels},
	titleaddon = {{arXiv}.org},
	author = {Cordeiro, Filipe R. and Carneiro, Gustavo},
	date = {2020-12-05},
	langid = {english},
}

@inproceedings{berthon_confidence_2021,
	title = {Confidence Scores Make Instance-dependent Label-noise Learning Possible},
	issn = {2640-3498},
	eventtitle = {International Conference on Machine Learning},
	pages = {825--836},
	booktitle = {Proceedings of the 38th International Conference on Machine Learning},
	publisher = {{PMLR}},
	author = {Berthon, Antonin and Han, Bo and Niu, Gang and Liu, Tongliang and Sugiyama, Masashi},
	date = {2021-07-01},
	langid = {english},
}

@article{ghosh_making_2015,
	title = {Making risk minimization tolerant to label noise},
	volume = {160},
	issn = {0925-2312},
	doi = {10.1016/j.neucom.2014.09.081},
	pages = {93--107},
	journaltitle = {Neurocomputing},
	shortjournal = {Neurocomputing},
	author = {Ghosh, Aritra and Manwani, Naresh and Sastry, P. S.},
	date = {2015-07-21},
}

@article{philps_uncertainty_2025,
	title = {Uncertainty quantification for White Matter Hyperintensity segmentation detects silent failures and improves automated Fazekas quantification},
	volume = {105},
	issn = {1361-8415},
	doi = {10.1016/j.media.2025.103697},
	pages = {103697},
	journaltitle = {Medical Image Analysis},
	shortjournal = {Medical Image Analysis},
	author = {Philps, Ben and Valdés Hernández, Maria del C. and Qin, Chen and Clancy, Una and Sakka, Eleni and Muñoz Maniega, Susana and Bastin, Mark E. and Jochems, Angela C. C. and Wardlaw, Joanna M. and Bernabeu, Miguel O.},
	urldate = {2026-06-25},
	date = {2025-10-01},
}

@article{kuwabara_artificial_2024,
	title = {Artificial intelligence for volumetric measurement of cerebral white matter hyperintensities on thick-slice fluid-attenuated inversion recovery ({FLAIR}) magnetic resonance images from multiple centers},
	volume = {14},
	rights = {2024 The Author(s)},
	issn = {2045-2322},
	doi = {10.1038/s41598-024-60789-x},
	pages = {10104},
	number = {1},
	journaltitle = {Scientific Reports},
	shortjournal = {Sci Rep},
	publisher = {Nature Publishing Group},
	author = {Kuwabara, Masashi and Ikawa, Fusao and Nakazawa, Shinji and Koshino, Saori and Ishii, Daizo and Kondo, Hiroshi and Hara, Takeshi and Maeda, Yuyo and Sato, Ryo and Kaneko, Taiki and Maeyama, Shiyuki and Shimahara, Yuki and Horie, Nobutaka},
	urldate = {2025-07-02},
	date = {2024-05-02},
	langid = {english},
}

@article{avants_advanced_nodate,
	title = {Advanced Normalization Tools ({ANTS})},
	author = {Avants, Brian B and Tustison, Nick and Johnson, Hans},
	langid = {english},
}

@article{debette_clinical_2010,
	title = {The clinical importance of white matter hyperintensities on brain magnetic resonance imaging: systematic review and meta-analysis},
	volume = {341},
	issn = {1756-1833},
	doi = {10.1136/bmj.c3666},
	shorttitle = {The clinical importance of white matter hyperintensities on brain magnetic resonance imaging},
	pages = {c3666},
	journaltitle = {{BMJ} (Clinical research ed.)},
	shortjournal = {{BMJ}},
	author = {Debette, Stéphanie and Markus, H. S.},
	date = {2010-07-26},
	pmid = {20660506},
	pmcid = {PMC2910261},
}

@article{haller_brain_2013,
	title = {Do brain T2/{FLAIR} white matter hyperintensities correspond to myelin loss in normal aging? A radiologic-neuropathologic correlation study},
	volume = {1},
	issn = {2051-5960},
	doi = {10.1186/2051-5960-1-14},
	shorttitle = {Do brain T2/{FLAIR} white matter hyperintensities correspond to myelin loss in normal aging?},
	pages = {14},
	journaltitle = {Acta Neuropathologica Communications},
	shortjournal = {Acta Neuropathol Commun},
	author = {Haller, Sven and Kövari, Enikö and Herrmann, François R. and Cuvinciuc, Victor and Tomm, Ann-Marie and Zulian, Gilbert B. and Lovblad, Karl-Olof and Giannakopoulos, Panteleimon and Bouras, Constantin},
	date = {2013-05-09},
	pmid = {24252608},
	pmcid = {PMC3893472},
}

@article{fazekas_mr_1987,
	title = {{MR} signal abnormalities at 1.5 T in Alzheimer's dementia and normal aging},
	volume = {149},
	issn = {0361-803X},
	doi = {10.2214/ajr.149.2.351},
	pages = {351--356},
	number = {2},
	journaltitle = {{AJR}. American journal of roentgenology},
	shortjournal = {{AJR} Am J Roentgenol},
	author = {Fazekas, F. and Chawluk, J. B. and Alavi, A. and Hurtig, H. I. and Zimmerman, R. A.},
	date = {1987-08},
	pmid = {3496763},
}

@article{kim_classification_2008,
	title = {Classification of white matter lesions on magnetic resonance imaging in elderly persons},
	volume = {64},
	issn = {1873-2402},
	doi = {10.1016/j.biopsych.2008.03.024},
	pages = {273--280},
	number = {4},
	journaltitle = {Biological Psychiatry},
	shortjournal = {Biol Psychiatry},
	author = {Kim, Ki Woong and {MacFall}, James R. and Payne, Martha E.},
	date = {2008-08-15},
	pmid = {18471801},
	pmcid = {PMC2593803},
}

@article{bolandzadeh_association_2012,
	title = {The association between cognitive function and white matter lesion location in older adults: a systematic review},
	volume = {12},
	issn = {1471-2377},
	doi = {10.1186/1471-2377-12-126},
	shorttitle = {The association between cognitive function and white matter lesion location in older adults},
	pages = {126},
	journaltitle = {{BMC} neurology},
	shortjournal = {{BMC} Neurol},
	author = {Bolandzadeh, Niousha and Davis, Jennifer C. and Tam, Roger and Handy, Todd C. and Liu-Ambrose, Teresa},
	date = {2012-10-30},
	pmid = {23110387},
	pmcid = {PMC3522005},
}

@article{griffanti_classification_2018,
	title = {Classification and characterization of periventricular and deep white matter hyperintensities on {MRI}: A study in older adults},
	volume = {170},
	issn = {1095-9572},
	doi = {10.1016/j.neuroimage.2017.03.024},
	shorttitle = {Classification and characterization of periventricular and deep white matter hyperintensities on {MRI}},
	pages = {174--181},
	journaltitle = {{NeuroImage}},
	shortjournal = {Neuroimage},
	author = {Griffanti, Ludovica and Jenkinson, Mark and Suri, Sana and Zsoldos, Enikő and Mahmood, Abda and Filippini, Nicola and Sexton, Claire E. and Topiwala, Anya and Allan, Charlotte and Kivimäki, Mika and Singh-Manoux, Archana and Ebmeier, Klaus P. and Mackay, Clare E. and Zamboni, Giovanna},
	date = {2018-04-15},
	pmid = {28315460},
}

@article{krishnan_relationship_2006,
	title = {Relationship between periventricular and deep white matter lesions and depressive symptoms in older people. The {LADIS} Study},
	volume = {21},
	issn = {0885-6230},
	doi = {10.1002/gps.1596},
	pages = {983--989},
	number = {10},
	journaltitle = {International Journal of Geriatric Psychiatry},
	shortjournal = {Int J Geriatr Psychiatry},
	author = {Krishnan, Mani S. and O'Brien, John T. and Firbank, Michael J. and Pantoni, Leonardo and Carlucci, Giovanna and Erkinjuntti, Timo and Wallin, Anders and Wahlund, Lars-Olof and Scheltens, Philip and van Straaten, Elisabeth C. W. and Inzitari, Domenico and {LADIS Group}},
	date = {2006-10},
	pmid = {16955428},
}

@article{roseborough_microvessel_2022,
	title = {Microvessel stenosis, enlarged perivascular spaces, and fibrinogen deposition are associated with ischemic periventricular white matter hyperintensities},
	volume = {32},
	issn = {1750-3639},
	doi = {10.1111/bpa.13017},
	pages = {e13017},
	number = {1},
	journaltitle = {Brain Pathology (Zurich, Switzerland)},
	shortjournal = {Brain Pathol},
	author = {Roseborough, Austyn D. and Rasheed, Berk and Jung, Youngkyung and Nishimura, Kevin and Pinsky, William and Langdon, Kristopher D. and Hammond, Robert and Pasternak, Stephen H. and Khan, Ali R. and Whitehead, Shawn N.},
	date = {2022-01},
	pmid = {34538024},
	pmcid = {PMC8713528},
}

@article{lampe_visceral_2019,
	title = {Visceral obesity relates to deep white matter hyperintensities via inflammation},
	volume = {85},
	issn = {1531-8249},
	doi = {10.1002/ana.25396},
	pages = {194--203},
	number = {2},
	journaltitle = {Annals of Neurology},
	shortjournal = {Ann Neurol},
	author = {Lampe, Leonie and Zhang, Rui and Beyer, Frauke and Huhn, Sebastian and Kharabian Masouleh, Shahrzad and Preusser, Sven and Bazin, Pierre-Louis and Schroeter, Matthias L. and Villringer, Arno and Witte, A. Veronica},
	date = {2019-02},
	pmid = {30556596},
	pmcid = {PMC6590485},
}

@article{veldsman_spatial_2020,
	title = {Spatial distribution and cognitive impact of cerebrovascular risk-related white matter hyperintensities},
	volume = {28},
	issn = {2213-1582},
	doi = {10.1016/j.nicl.2020.102405},
	pages = {102405},
	journaltitle = {{NeuroImage}. Clinical},
	shortjournal = {Neuroimage Clin},
	author = {Veldsman, Michele and Kindalova, Petya and Husain, Masud and Kosmidis, Ioannis and Nichols, Thomas E.},
	date = {2020},
	pmid = {32971464},
	pmcid = {PMC7511743},
}

@article{wardlaw_white_2004,
	title = {White matter hyperintensities and rating scales-observer reliability varies with lesion load},
	volume = {251},
	issn = {0340-5354},
	doi = {10.1007/s00415-004-0371-x},
	pages = {584--590},
	number = {5},
	journaltitle = {Journal of Neurology},
	shortjournal = {J Neurol},
	author = {Wardlaw, Joanna M. and Ferguson, Karen J. and Graham, Catriona},
	date = {2004-05},
	pmid = {15164192},
}

@article{haughey_assessment_2025,
	title = {Assessment of Inter-Reader Reliability of Fazekas Scoring on Magnetic Resonance Imaging of the Brain in Adult Patients with Sickle Cell Disease},
	volume = {15},
	issn = {2075-4418},
	doi = {10.3390/diagnostics15070857},
	pages = {857},
	number = {7},
	journaltitle = {Diagnostics (Basel, Switzerland)},
	shortjournal = {Diagnostics (Basel)},
	author = {Haughey, Aoife M. and O'Cearbhaill, Roisin M. and Forté, Stephanie and Schaafsma, Joanna D. and Kuo, Kevin H. M. and Padilha, Igor Gomes},
	date = {2025-03-27},
	pmid = {40218207},
	pmcid = {PMC11988997},
}

@article{balakrishnan_automatic_2021,
	title = {Automatic segmentation of white matter hyperintensities from brain magnetic resonance images in the era of deep learning and big data - A systematic review},
	volume = {88},
	issn = {1879-0771},
	doi = {10.1016/j.compmedimag.2021.101867},
	pages = {101867},
	journaltitle = {Computerized Medical Imaging and Graphics: The Official Journal of the Computerized Medical Imaging Society},
	shortjournal = {Comput Med Imaging Graph},
	author = {Balakrishnan, Ramya and Valdés Hernández, Maria Del C. and Farrall, Andrew J.},
	date = {2021-03},
	pmid = {33508567},
}

@article{umapathy_stacked_2021,
	title = {A Stacked Generalization of 3D Orthogonal Deep Learning Convolutional Neural Networks for Improved Detection of White Matter Hyperintensities in 3D {FLAIR} Images},
	volume = {42},
	issn = {1936-959X},
	doi = {10.3174/ajnr.A6970},
	pages = {639--647},
	number = {4},
	journaltitle = {{AJNR}. American journal of neuroradiology},
	shortjournal = {{AJNR} Am J Neuroradiol},
	author = {Umapathy, L. and Perez-Carrillo, G. G. and Keerthivasan, M. B. and Rosado-Toro, J. A. and Altbach, M. I. and Winegar, B. and Weinkauf, C. and Bilgin, A. and {Alzheimer’s Disease Neuroimaging Initiative}},
	date = {2021-04},
	pmid = {33574101},
	pmcid = {PMC8040994},
}

@article{heinen_performance_2019,
	title = {Performance of five automated white matter hyperintensity segmentation methods in a multicenter dataset},
	volume = {9},
	issn = {2045-2322},
	doi = {10.1038/s41598-019-52966-0},
	pages = {16742},
	number = {1},
	journaltitle = {Scientific Reports},
	shortjournal = {Sci Rep},
	author = {Heinen, Rutger and Steenwijk, Martijn D. and Barkhof, Frederik and Biesbroek, J. Matthijs and van der Flier, Wiesje M. and Kuijf, Hugo J. and Prins, Niels D. and Vrenken, Hugo and Biessels, Geert Jan and de Bresser, Jeroen and {TRACE-VCI study group}},
	date = {2019-11-14},
	pmid = {31727919},
	pmcid = {PMC6856351},
}

@article{melazzini_white_2021,
	title = {White Matter Hyperintensities Quantification in Healthy Adults: A Systematic Review and Meta-Analysis},
	volume = {53},
	issn = {1522-2586},
	doi = {10.1002/jmri.27479},
	shorttitle = {White Matter Hyperintensities Quantification in Healthy Adults},
	pages = {1732--1743},
	number = {6},
	journaltitle = {Journal of magnetic resonance imaging: {JMRI}},
	shortjournal = {J Magn Reson Imaging},
	author = {Melazzini, Luca and Vitali, Paolo and Olivieri, Emanuele and Bolchini, Marco and Zanardo, Moreno and Savoldi, Filippo and Di Leo, Giovanni and Griffanti, Ludovica and Baselli, Giuseppe and Sardanelli, Francesco and Codari, Marina},
	date = {2021-06},
	pmid = {33345393},
}

@article{karimi_deep_2020,
	title = {Deep learning with noisy labels: Exploring techniques and remedies in medical image analysis},
	volume = {65},
	issn = {1361-8423},
	doi = {10.1016/j.media.2020.101759},
	shorttitle = {Deep learning with noisy labels},
	pages = {101759},
	journaltitle = {Medical Image Analysis},
	shortjournal = {Med Image Anal},
	author = {Karimi, Davood and Dou, Haoran and Warfield, Simon K. and Gholipour, Ali},
	date = {2020-10},
	pmid = {32623277},
	pmcid = {PMC7484266},
}

@article{gisev_interrater_2013,
	title = {Interrater agreement and interrater reliability: key concepts, approaches, and applications},
	volume = {9},
	issn = {1934-8150},
	doi = {10.1016/j.sapharm.2012.04.004},
	shorttitle = {Interrater agreement and interrater reliability},
	pages = {330--338},
	number = {3},
	journaltitle = {Research in social \& administrative pharmacy: {RSAP}},
	shortjournal = {Res Social Adm Pharm},
	author = {Gisev, Natasa and Bell, J. Simon and Chen, Timothy F.},
	date = {2013},
	pmid = {22695215},
}

@article{zhang_generalized_2018,
	title = {Generalized Cross Entropy Loss for Training Deep Neural Networks with Noisy Labels},
	volume = {32},
	issn = {1049-5258},
	pages = {8792--8802},
	journaltitle = {Advances in Neural Information Processing Systems},
	shortjournal = {Adv Neural Inf Process Syst},
	author = {Zhang, Zhilu and Sabuncu, Mert R.},
	date = {2018-12},
	pmid = {39839708},
	pmcid = {PMC11747755},
}

@inproceedings{castagnos_simple_2022,
	location = {Gyeongju, Republic of Korea},
	title = {A Simple Log-based Loss Function for Ordinal Text Classification},
	url = {https://aclanthology.org/2022.coling-1.407/},
	eventtitle = {{COLING} 2022},
	pages = {4604--4609},
	booktitle = {Proceedings of the 29th International Conference on Computational Linguistics},
	publisher = {International Committee on Computational Linguistics},
	author = {Castagnos, François and Mihelich, Martin and Dognin, Charles},
	editor = {Calzolari, Nicoletta and Huang, Chu-Ren and Kim, Hansaem and Pustejovsky, James and Wanner, Leo and Choi, Key-Sun and Ryu, Pum-Mo and Chen, Hsin-Hsi and Donatelli, Lucia and Ji, Heng and Kurohashi, Sadao and Paggio, Patrizia and Xue, Nianwen and Kim, Seokhwan and Hahm, Younggyun and He, Zhong and Lee, Tony Kyungil and Santus, Enrico and Bond, Francis and Na, Seung-Hoon},
	urldate = {2025-06-26},
	date = {2022-10},
}

@inproceedings{wei_mitigating_2023,
	title = {Mitigating Memorization of Noisy Labels by Clipping the Model Prediction},
	url = {https://proceedings.mlr.press/v202/wei23e.html},
	eventtitle = {International Conference on Machine Learning},
	pages = {36868--36886},
	booktitle = {Proceedings of the 40th International Conference on Machine Learning},
	publisher = {{PMLR}},
	author = {Wei, Hongxin and Zhuang, Huiping and Xie, Renchunzi and Feng, Lei and Niu, Gang and An, Bo and Li, Yixuan},
	urldate = {2025-06-26},
	date = {2023-07-03},
	langid = {english},
	note = {{ISSN}: 2640-3498},
}

@article{chicco_advantages_2020,
	title = {The advantages of the Matthews correlation coefficient ({MCC}) over F1 score and accuracy in binary classification evaluation},
	volume = {21},
	issn = {1471-2164},
	doi = {10.1186/s12864-019-6413-7},
	pages = {6},
	number = {1},
	journaltitle = {{BMC} genomics},
	shortjournal = {{BMC} Genomics},
	author = {Chicco, Davide and Jurman, Giuseppe},
	date = {2020-01-02},
	pmid = {31898477},
	pmcid = {PMC6941312},
}

@article{maaten_visualizing_2008,
	title = {Visualizing Data using t-{SNE}},
	volume = {9},
	issn = {1533-7928},
	url = {http://jmlr.org/papers/v9/vandermaaten08a.html},
	pages = {2579--2605},
	number = {86},
	journaltitle = {Journal of Machine Learning Research},
	author = {Maaten, Laurens van der and Hinton, Geoffrey},
	urldate = {2025-06-26},
	date = {2008},
}

@inproceedings{gal_dropout_2016,
	title = {Dropout as a Bayesian Approximation: Representing Model Uncertainty in Deep Learning},
	url = {https://proceedings.mlr.press/v48/gal16.html},
	shorttitle = {Dropout as a Bayesian Approximation},
	eventtitle = {International Conference on Machine Learning},
	pages = {1050--1059},
	booktitle = {Proceedings of The 33rd International Conference on Machine Learning},
	publisher = {{PMLR}},
	author = {Gal, Yarin and Ghahramani, Zoubin},
	urldate = {2025-06-26},
	date = {2016-06-11},
	langid = {english},
	note = {{ISSN}: 1938-7228},
}

@incollection{montavon_layer-wise_2019,
	location = {Cham},
	title = {Layer-Wise Relevance Propagation: An Overview},
	isbn = {978-3-030-28954-6},
	url = {https://doi.org/10.1007/978-3-030-28954-6_10},
	shorttitle = {Layer-Wise Relevance Propagation},
	pages = {193--209},
	booktitle = {Explainable {AI}: Interpreting, Explaining and Visualizing Deep Learning},
	publisher = {Springer International Publishing},
	author = {Montavon, Grégoire and Binder, Alexander and Lapuschkin, Sebastian and Samek, Wojciech and Müller, Klaus-Robert},
	editor = {Samek, Wojciech and Montavon, Grégoire and Vedaldi, Andrea and Hansen, Lars Kai and Müller, Klaus-Robert},
	urldate = {2025-06-26},
	date = {2019},
	langid = {english},
	doi = {10.1007/978-3-030-28954-6_10},
}

@inproceedings{kendall_what_2017,
	location = {Red Hook, {NY}, {USA}},
	title = {What uncertainties do we need in Bayesian deep learning for computer vision?},
	isbn = {978-1-5108-6096-4},
	series = {{NIPS}'17},
	pages = {5580--5590},
	booktitle = {Proceedings of the 31st International Conference on Neural Information Processing Systems},
	publisher = {Curran Associates Inc.},
	author = {Kendall, Alex and Gal, Yarin},
	urldate = {2025-06-26},
	date = {2017-12-04},
}

@article{mccreary_calgary_2020,
	title = {Calgary Normative Study: design of a prospective longitudinal study to characterise potential quantitative {MR} biomarkers of neurodegeneration over the adult lifespan},
	volume = {10},
	issn = {2044-6055},
	doi = {10.1136/bmjopen-2020-038120},
	shorttitle = {Calgary Normative Study},
	pages = {e038120},
	number = {8},
	journaltitle = {{BMJ} open},
	shortjournal = {{BMJ} Open},
	author = {{McCreary}, Cheryl R. and Salluzzi, Marina and Andersen, Linda B. and Gobbi, David and Lauzon, Louis and Saad, Feryal and Smith, Eric E. and Frayne, Richard},
	date = {2020-08-13},
	pmid = {32792445},
	pmcid = {PMC7430487},
}

@article{tustison_antsx_2021,
	title = {The {ANTsX} ecosystem for quantitative biological and medical imaging},
	volume = {11},
	rights = {2021 The Author(s)},
	issn = {2045-2322},
	url = {https://www.nature.com/articles/s41598-021-87564-6},
	doi = {10.1038/s41598-021-87564-6},
	pages = {9068},
	number = {1},
	journaltitle = {Scientific Reports},
	shortjournal = {Sci Rep},
	author = {Tustison, Nicholas J. and Cook, Philip A. and Holbrook, Andrew J. and Johnson, Hans J. and Muschelli, John and Devenyi, Gabriel A. and Duda, Jeffrey T. and Das, Sandhitsu R. and Cullen, Nicholas C. and Gillen, Daniel L. and Yassa, Michael A. and Stone, James R. and Gee, James C. and Avants, Brian B.},
	urldate = {2025-06-26},
	date = {2021-04-27},
	langid = {english},
	note = {Publisher: Nature Publishing Group},
}

@article{smith_advances_2004,
	title = {Advances in functional and structural {MR} image analysis and implementation as {FSL}},
	volume = {23 Suppl 1},
	issn = {1053-8119},
	doi = {10.1016/j.neuroimage.2004.07.051},
	pages = {S208--219},
	journaltitle = {{NeuroImage}},
	shortjournal = {Neuroimage},
	author = {Smith, Stephen M. and Jenkinson, Mark and Woolrich, Mark W. and Beckmann, Christian F. and Behrens, Timothy E. J. and Johansen-Berg, Heidi and Bannister, Peter R. and De Luca, Marilena and Drobnjak, Ivana and Flitney, David E. and Niazy, Rami K. and Saunders, James and Vickers, John and Zhang, Yongyue and De Stefano, Nicola and Brady, J. Michael and Matthews, Paul M.},
	date = {2004},
	pmid = {15501092},
}

@article{song_learning_2023,
	title = {Learning From Noisy Labels With Deep Neural Networks: A Survey},
	volume = {34},
	issn = {2162-2388},
	doi = {10.1109/TNNLS.2022.3152527},
	shorttitle = {Learning From Noisy Labels With Deep Neural Networks},
	pages = {8135--8153},
	number = {11},
	journaltitle = {{IEEE} transactions on neural networks and learning systems},
	shortjournal = {{IEEE} Trans Neural Netw Learn Syst},
	author = {Song, Hwanjun and Kim, Minseok and Park, Dongmin and Shin, Yooju and Lee, Jae-Gil},
	date = {2023-11},
	pmid = {35254993},
}

@article{liu_deep_2020,
	title = {Deep convolutional neural network for accurate segmentation and quantification of white matter hyperintensities},
	volume = {384},
	issn = {0925-2312},
	url = {https://www.sciencedirect.com/science/article/pii/S092523121931759X},
	doi = {10.1016/j.neucom.2019.12.050},
	pages = {231--242},
	journaltitle = {Neurocomputing},
	shortjournal = {Neurocomputing},
	author = {Liu, Liangliang and Chen, Shaowu and Zhu, Xiaofeng and Zhao, Xing-Ming and Wu, Fang-Xiang and Wang, Jianxin},
	urldate = {2025-06-26},
	date = {2020-04-07},
}

@article{duarte_segmenting_2023,
	title = {Segmenting white matter hyperintensities in brain magnetic resonance images using convolution neural networks},
	volume = {175},
	issn = {0167-8655},
	url = {https://www.sciencedirect.com/science/article/pii/S0167865523002179},
	doi = {10.1016/j.patrec.2023.07.014},
	pages = {90--94},
	journaltitle = {Pattern Recognition Letters},
	shortjournal = {Pattern Recognition Letters},
	author = {Duarte, Kauê T. N. and Gobbi, David G. and Sidhu, Abhijot S. and {McCreary}, Cheryl R. and Saad, Feryal and Camicioli, Richard and Smith, Eric E. and Frayne, Richard},
	urldate = {2025-06-26},
	date = {2023-11-01},
}

@article{frenay_classification_2014,
	title = {Classification in the presence of label noise: a survey},
	volume = {25},
	issn = {2162-2388},
	doi = {10.1109/TNNLS.2013.2292894},
	shorttitle = {Classification in the presence of label noise},
	pages = {845--869},
	number = {5},
	journaltitle = {{IEEE} transactions on neural networks and learning systems},
	shortjournal = {{IEEE} Trans Neural Netw Learn Syst},
	author = {Frénay, Benoît and Verleysen, Michel},
	date = {2014-05},
	pmid = {24808033},
}

@article{bahrani_post-acquisition_2019,
	title = {Post-acquisition processing confounds in brain volumetric quantification of white matter hyperintensities},
	volume = {327},
	issn = {1872-678X},
	doi = {10.1016/j.jneumeth.2019.108391},
	pages = {108391},
	journaltitle = {Journal of Neuroscience Methods},
	shortjournal = {J Neurosci Methods},
	author = {Bahrani, Ahmed A. and Al-Janabi, Omar M. and Abner, Erin L. and Bardach, Shoshana H. and Kryscio, Richard J. and Wilcock, Donna M. and Smith, Charles D. and Jicha, Gregory A.},
	date = {2019-11-01},
	pmid = {31408649},
	pmcid = {PMC6746343},
}

@article{rieu_fully_2023,
	title = {A Fully Automated Visual Grading System for White Matter Hyperintensities of T2-Fluid Attenuated Inversion Recovery Magnetic Resonance Imaging},
	volume = {22},
	issn = {0219-6352},
	doi = {10.31083/j.jin2203057},
	pages = {57},
	number = {3},
	journaltitle = {Journal of Integrative Neuroscience},
	shortjournal = {J Integr Neurosci},
	author = {Rieu, {ZunHyan} and Kim, Regina Ey and Lee, Minho and Kim, Hye Weon and Kim, Donghyeon and Yong, {JeongHyun} and Kim, {JiMin} and Lee, {MinKyoung} and Lim, Hyunkook and Kim, {JeeYoung}},
	date = {2023-05-06},
	pmid = {37258435},
}

@article{joo_diagnostic_2022,
	title = {Diagnostic performance of deep learning-based automatic white matter hyperintensity segmentation for classification of the Fazekas scale and differentiation of subcortical vascular dementia},
	volume = {17},
	issn = {1932-6203},
	doi = {10.1371/journal.pone.0274562},
	pages = {e0274562},
	number = {9},
	journaltitle = {{PloS} One},
	shortjournal = {{PLoS} One},
	author = {Joo, Leehi and Shim, Woo Hyun and Suh, Chong Hyun and Lim, Su Jin and Heo, Hwon and Kim, Woo Seok and Hong, Eunpyeong and Lee, Dongsoo and Sung, Jinkyeong and Lim, Jae-Sung and Lee, Jae-Hong and Kim, Sang Joon},
	date = {2022},
	pmid = {36107961},
	pmcid = {PMC9477348},
}

@article{samek_evaluating_2017,
	title = {Evaluating the Visualization of What a Deep Neural Network Has Learned},
	volume = {28},
	rights = {https://ieeexplore.ieee.org/Xplorehelp/downloads/license-information/{IEEE}.html},
	issn = {2162-237X, 2162-2388},
	url = {https://ieeexplore.ieee.org/document/7552539/},
	doi = {10.1109/TNNLS.2016.2599820},
	pages = {2660--2673},
	number = {11},
	journaltitle = {{IEEE} Transactions on Neural Networks and Learning Systems},
	shortjournal = {{IEEE} Trans. Neural Netw. Learning Syst.},
	author = {Samek, Wojciech and Binder, Alexander and Montavon, Gregoire and Lapuschkin, Sebastian and Muller, Klaus-Robert},
	urldate = {2025-06-26},
	date = {2017-11},
}

@article{wardlaw_neuroimaging_2013,
	title = {Neuroimaging standards for research into small vessel disease and its contribution to ageing and neurodegeneration},
	volume = {12},
	issn = {1474-4465},
	doi = {10.1016/S1474-4422(13)70124-8},
	pages = {822--838},
	number = {8},
	journaltitle = {The Lancet. Neurology},
	shortjournal = {Lancet Neurol},
	author = {Wardlaw, Joanna M. and Smith, Eric E. and Biessels, Geert J. and Cordonnier, Charlotte and Fazekas, Franz and Frayne, Richard and Lindley, Richard I. and O'Brien, John T. and Barkhof, Frederik and Benavente, Oscar R. and Black, Sandra E. and Brayne, Carol and Breteler, Monique and Chabriat, Hugues and Decarli, Charles and de Leeuw, Frank-Erik and Doubal, Fergus and Duering, Marco and Fox, Nick C. and Greenberg, Steven and Hachinski, Vladimir and Kilimann, Ingo and Mok, Vincent and Oostenbrugge, Robert van and Pantoni, Leonardo and Speck, Oliver and Stephan, Blossom C. M. and Teipel, Stefan and Viswanathan, Anand and Werring, David and Chen, Christopher and Smith, Colin and van Buchem, Mark and Norrving, Bo and Gorelick, Philip B. and Dichgans, Martin and {STandards for ReportIng Vascular changes on nEuroimaging (STRIVE v1)}},
	date = {2013-08},
	pmid = {23867200},
	pmcid = {PMC3714437},
}

@article{huang_uncertainty-aware_2022,
	title = {Uncertainty-Aware Learning against Label Noise on Imbalanced Datasets},
	volume = {36},
	rights = {Copyright (c) 2022 Association for the Advancement of Artificial Intelligence},
	issn = {2374-3468},
	url = {https://ojs.aaai.org/index.php/AAAI/article/view/20654},
	doi = {10.1609/aaai.v36i6.20654},
	pages = {6960--6969},
	number = {6},
	journaltitle = {Proceedings of the {AAAI} Conference on Artificial Intelligence},
	author = {Huang, Yingsong and Bai, Bing and Zhao, Shengwei and Bai, Kun and Wang, Fei},
	urldate = {2025-06-26},
	date = {2022-06-28},
	langid = {english},
	note = {Number: 6},
}

@article{xu_usdnl_2023,
	title = {{USDNL}: Uncertainty-Based Single Dropout in Noisy Label Learning},
	volume = {37},
	rights = {Copyright (c) 2023 Association for the Advancement of Artificial Intelligence},
	issn = {2374-3468},
	url = {https://ojs.aaai.org/index.php/AAAI/article/view/26264},
	doi = {10.1609/aaai.v37i9.26264},
	shorttitle = {{USDNL}},
	pages = {10648--10656},
	number = {9},
	journaltitle = {Proceedings of the {AAAI} Conference on Artificial Intelligence},
	author = {Xu, Yuanzhuo and Niu, Xiaoguang and Yang, Jie and Drew, Steve and Zhou, Jiayu and Chen, Ruizhi},
	urldate = {2025-06-26},
	date = {2023-06-26},
	langid = {english},
	note = {Number: 9},
}

@article{ju_improving_2022,
	title = {Improving Medical Images Classification With Label Noise Using Dual-Uncertainty Estimation},
	volume = {41},
	issn = {1558-254X},
	url = {https://ieeexplore.ieee.org/abstract/document/9674886},
	doi = {10.1109/TMI.2022.3141425},
	pages = {1533--1546},
	number = {6},
	journaltitle = {{IEEE} Transactions on Medical Imaging},
	author = {Ju, Lie and Wang, Xin and Wang, Lin and Mahapatra, Dwarikanath and Zhao, Xin and Zhou, Quan and Liu, Tongliang and Ge, Zongyuan},
	urldate = {2025-06-26},
	date = {2022-06},
}

@inproceedings{schmid_quantifying_2025,
	title = {Quantifying White Matter Hyperintensities: Predicting Periventricular Fazekas Scores with Uncertainty Estimation},
	url = {https://ieeexplore.ieee.org/abstract/document/10981010},
	doi = {10.1109/ISBI60581.2025.10981010},
	shorttitle = {Quantifying White Matter Hyperintensities},
	eventtitle = {2025 {IEEE} 22nd International Symposium on Biomedical Imaging ({ISBI})},
	pages = {1--5},
	booktitle = {2025 {IEEE} 22nd International Symposium on Biomedical Imaging ({ISBI})},
	author = {Schmid, Susanne and Shamaei, Amirmohammed and Souza, Roberto and Frayne, Richard},
	urldate = {2025-06-27},
	date = {2025-04},
	note = {{ISSN}: 1945-8452},
}

@article{salahuddin_transparency_2022,
	title = {Transparency of deep neural networks for medical image analysis: A review of interpretability methods},
	volume = {140},
	issn = {0010-4825},
	url = {https://www.sciencedirect.com/science/article/pii/S0010482521009057},
	doi = {10.1016/j.compbiomed.2021.105111},
	shorttitle = {Transparency of deep neural networks for medical image analysis},
	pages = {105111},
	journaltitle = {Computers in Biology and Medicine},
	shortjournal = {Computers in Biology and Medicine},
	author = {Salahuddin, Zohaib and Woodruff, Henry C. and Chatterjee, Avishek and Lambin, Philippe},
	urldate = {2025-06-27},
	date = {2022-01-01},
}

@article{shi_survey_2024,
	title = {A survey of label-noise deep learning for medical image analysis},
	volume = {95},
	issn = {1361-8415},
	doi = {10.1016/j.media.2024.103166},
	pages = {103166},
	journaltitle = {Medical Image Analysis},
	shortjournal = {Medical Image Analysis},
	author = {Shi, Jialin and Zhang, Kailai and Guo, Chenyi and Yang, Youquan and Xu, Yali and Wu, Ji},
	urldate = {2026-07-06},
	date = {2024-07-01},
}

@article{sole-guardia_impact_2025,
	title = {Impact of hypertension on cerebral small vessel disease: A post-mortem study of microvascular pathology from normal-appearing white matter into white matter hyperintensities},
	issn = {0271-678X},
	url = {https://doi.org/10.1177/0271678X251333256},
	doi = {10.1177/0271678X251333256},
	shorttitle = {Impact of hypertension on cerebral small vessel disease},
	pages = {0271678X251333256},
	journaltitle = {Journal of Cerebral Blood Flow \& Metabolism},
	shortjournal = {J Cereb Blood Flow Metab},
	author = {Solé-Guardia, Gemma and Janssen, Anne and Wolters, Rowan and Dohmen, Tren and Küsters, Benno and Claassen, Jurgen {AHR} and Leeuw, Frank-Erik de and Wiesmann, Maximilian and Gutierrez, Jose and Kiliaan, Amanda J},
	urldate = {2025-07-02},
	date = {2025-04-12},
	note = {Publisher: {SAGE} Publications Ltd {STM}},
}

@article{kruggel_impact_2010,
	title = {Impact of scanner hardware and imaging protocol on image quality and compartment volume precision in the {ADNI} cohort},
	volume = {49},
	issn = {1053-8119},
	url = {https://www.sciencedirect.com/science/article/pii/S1053811909011902},
	doi = {10.1016/j.neuroimage.2009.11.006},
	pages = {2123--2133},
	number = {3},
	journaltitle = {{NeuroImage}},
	shortjournal = {{NeuroImage}},
	author = {Kruggel, Frithjof and Turner, Jessica and Muftuler, L. Tugan},
	urldate = {2025-07-03},
	date = {2010-02-01},
}

@inproceedings{xia_part-dependent_2020,
	title = {Part-dependent Label Noise: Towards Instance-dependent Label Noise},
	volume = {33},
	url = {https://proceedings.neurips.cc/paper_files/paper/2020/hash/5607fe8879e4fd269e88387e8cb30b7e-Abstract.html},
	shorttitle = {Part-dependent Label Noise},
	pages = {7597--7610},
	booktitle = {Advances in Neural Information Processing Systems},
	publisher = {Curran Associates, Inc.},
	author = {Xia, Xiaobo and Liu, Tongliang and Han, Bo and Wang, Nannan and Gong, Mingming and Liu, Haifeng and Niu, Gang and Tao, Dacheng and Sugiyama, Masashi},
	urldate = {2025-07-04},
	date = {2020},
}

@article{liao_instance-dependent_2025,
	title = {Instance-dependent Label Distribution Estimation for Learning with Label Noise},
	volume = {133},
	issn = {1573-1405},
	url = {https://doi.org/10.1007/s11263-024-02299-x},
	doi = {10.1007/s11263-024-02299-x},
	pages = {2568--2580},
	number = {5},
	journaltitle = {International Journal of Computer Vision},
	shortjournal = {Int J Comput Vis},
	author = {Liao, Zehui and Hu, Shishuai and Xie, Yutong and Xia, Yong},
	urldate = {2025-07-04},
	date = {2025-05-01},
	langid = {english},
}

@article{chicco_matthews_2023,
	title = {The Matthews correlation coefficient ({MCC}) should replace the {ROC} {AUC} as the standard metric for assessing binary classification},
	volume = {16},
	issn = {1756-0381},
	url = {https://doi.org/10.1186/s13040-023-00322-4},
	doi = {10.1186/s13040-023-00322-4},
	pages = {4},
	number = {1},
	journaltitle = {{BioData} Mining},
	shortjournal = {{BioData} Mining},
	author = {Chicco, Davide and Jurman, Giuseppe},
	urldate = {2025-07-10},
	date = {2023-02-17},
	langid = {english},
}



\end{document}